\documentclass[letterpaper]{article} 
\usepackage{aaai2026}  
\usepackage{times}  
\usepackage{helvet}  
\usepackage{courier}  
\usepackage[hyphens]{url}  
\usepackage{graphicx} 
\usepackage{natbib}  
\usepackage{caption} 
\usepackage{algorithm}
\usepackage{algorithmic}

\usepackage{newfloat}
\usepackage{listings}
\DeclareCaptionStyle{ruled}{labelfont=normalfont,labelsep=colon,strut=off} 
\floatstyle{ruled}
\newfloat{listing}{tb}{lst}{}
\floatname{listing}{Listing}
\usepackage{graphicx}
\usepackage{amsmath}
\usepackage{amsfonts}
\usepackage{xcolor}
\usepackage{multirow}
\usepackage{rotating}
\usepackage{booktabs}

\def\qed{\space$\square$ \par \vspace{.15in}}
\def\hat{\widehat}

\newcommand{\bc}{\begin{center}}
\newcommand{\ec}{\end{center}}
\newcommand{\be}{\begin{equation}}
\newcommand{\ee}{\end{equation}}
\newcommand{\ba}{\begin{array}}
\newcommand{\ea}{\end{array}}
\newcommand{\bean}{\begin{eqnarray*}}
\newcommand{\eean}{\end{eqnarray*}}
\newcommand{\bea}{\begin{eqnarray}}
\newcommand{\eea}{\end{eqnarray}}
\newcommand{\ben}{\begin{enumerate}}
\newcommand{\een}{\end{enumerate}}
\newcommand{\bed}{\begin{itemize}}
\newcommand{\eed}{\end{itemize}}

\newtheorem{proposition}{\bf Proposition}
\newtheorem{remark}{\bf Remark}

\makeatletter
\def\copyright@on{F}
\def\showauthors@on{T}
\makeatother

\title{MIRRAMS: Learning Robust Tabular Models under Unseen Missingness Shifts}
\author{
    Jihye Lee\textsuperscript{\rm 1},
    Minseo Kang\textsuperscript{\rm 1},
    Dongha Kim\textsuperscript{\rm 1}\textsuperscript{\rm 2}\thanks{Corresponding author.}
}
\affiliations{
    \textsuperscript{\rm 1}Department of Statistics, Sungshin Women’s University\\
    \textsuperscript{\rm 2}Data Science Center, Sungshin Women’s University

}

\usepackage{bibentry}

\begin{document}

\maketitle

\begin{abstract}
The presence of missing values often reflects variations in data collection policies, which may shift across time or locations, even when the underlying feature distribution remains stable.
Such shifts in the missingness distribution between training and test inputs pose a significant challenge to achieving robust predictive performance.
In this study, we propose a novel deep learning framework designed to address this challenge, 
particularly in the common yet challenging scenario where the test-time dataset is \textit{unseen}.
We begin by introducing a set of mutual information-based conditions, called \textit{MI robustness conditions}, which guide the prediction model to extract label-relevant information. 
This promotes robustness against distributional shifts in missingness at test-time.
To enforce these conditions, we design simple yet effective loss terms that collectively define our final objective, called \textit{MIRRAMS}.
Importantly, our method does not rely on any specific missingness assumption such as MCAR, MAR, or MNAR, making it applicable to a broad range of scenarios.
Furthermore, it can naturally extend to cases where labels are also missing in training data, by generalizing the framework to a semi-supervised learning setting.
Extensive experiments across multiple benchmark tabular datasets demonstrate that MIRRAMS consistently outperforms existing state-of-the-art baselines and maintains stable performance under diverse missingness conditions. 
Moreover, it achieves superior performance even in fully observed settings, highlighting MIRRAMS as a powerful, off-the-shelf framework for general-purpose tabular learning.
\end{abstract}


\section{Introduction}

\label{sec:intro}
\paragraph{Missingness Shifts}
In real-world scenarios, it is quite common for the missing data distribution to differ between the training and test datasets. 
This often arises due to changes in data collection environments, population shifts, or operational procedures. 
For example, hospitals may record patient variables differently \citep{saez2020ehrtemporalvariability,zhou2023domain, stokes2025domain} and financial regulations can affect how information is reported \citep{qian2022managing}.
In addition, survey respondent demographics may change over time \citep{meterko2015response}, all of which can lead to shifts in the missing data distribution.

Such shifts in missing data distributions degrade model performance, as patterns learned from the observed training data may no longer be effective in predicting the test data \citep{gardner2023benchmarking}. 
This issue is particularly critical in tabular data, where each variable often carries distinct semantic meaning \citep{kim2023review}, making models more sensitive to missingness shifts than in unstructured domains such as image or text.
Although this is a practically important and pressing challenge, and despite the recent development of numerous deep-learning-based architectures for tabular data \citep{chen2023recontab,wu2024switchtab}, to the best of the authors' knowledge, there has been limited deep learning research specifically aimed at addressing this problem \citep{rockenschaub2024robust}.

It is known that when the test-time input data (i.e., target data) are accessible in advance, the problem can be framed as a form of domain adaptation \citep{DBLP:conf/cvpr/TzengHSD17,DBLP:conf/icml/XieZCC18}, and several recent studies have proposed methods under this setting \citep{ouyang2021maximum, zhou2023domain}. 
In this study, however, we consider a more general and challenging setting in which the test-time dataset is \textit{entirely unseen}, making existing domain adaptation methods inapplicable.

\paragraph{Overview of Our Study}
We propose a new deep learning framework for training prediction models that are \textit{robust} to \textit{unseen} missingness shifts, primarily designed for tabular data. 
To this end, we introduce a set of mutual information-based conditions, referred to as \textit{MI robustness conditions}, designed to guide the prediction model in extracting label-relevant information from inputs, regardless of missingness patterns in both the training and test data.
However, as the test data distribution is unavailable during training and computing mutual information is typically intractable, directly enforcing the MI robustness conditions becomes non-trivial.
	
\begin{figure}		
    \centering
    \includegraphics[width=\linewidth]{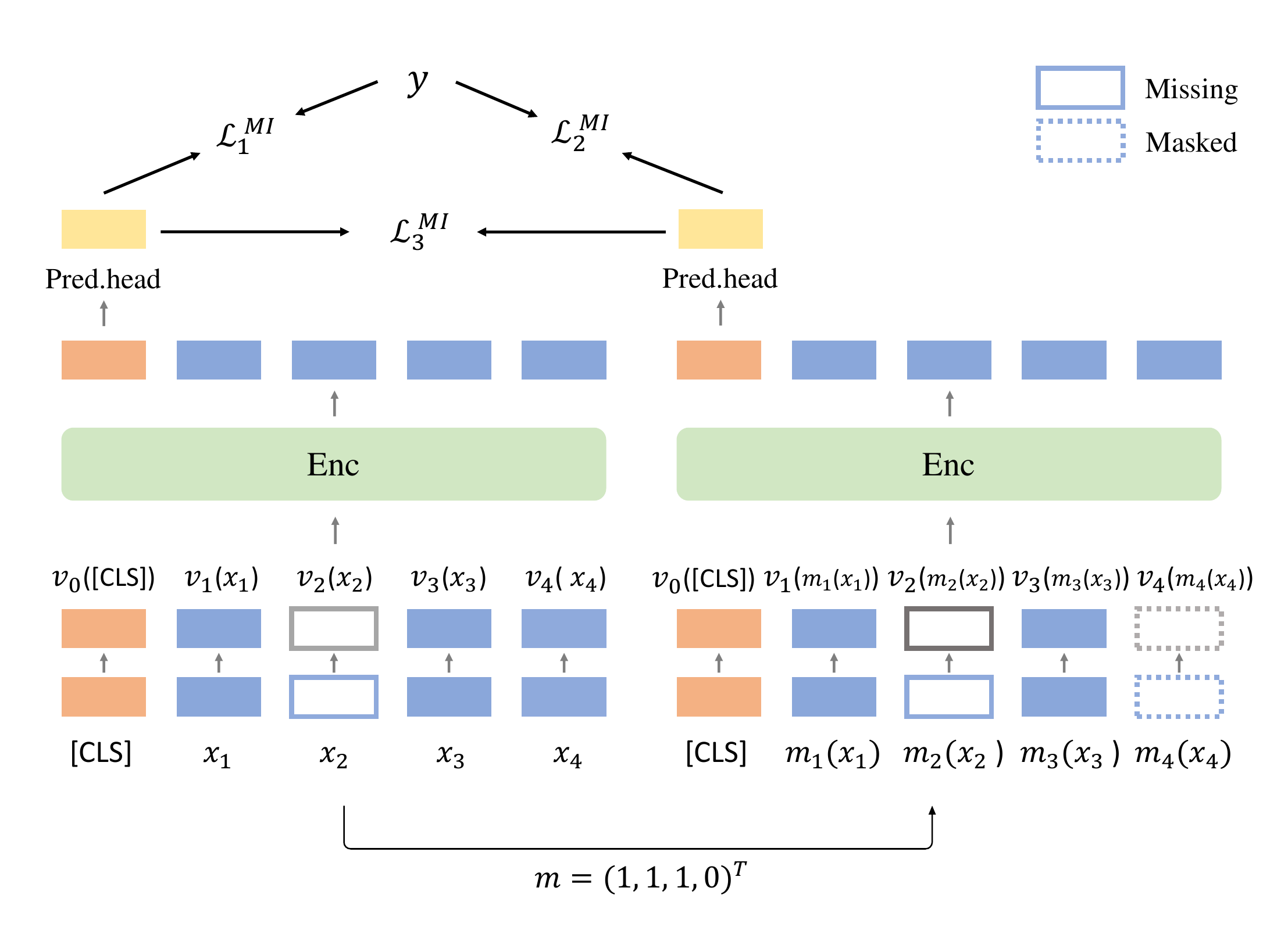}
    \caption{An illustration of the implementation of MIRRAMS on data with $p=4$. 
    }
    \label{fig:mirrams}
\end{figure}

To tackle these challenges, we introduce a set of novel techniques that are simple yet effective.
First, we apply additional masking to training data to \textit{realistically simulate unseen missingness patterns}, thereby better covering the potential support of the test-time missingness distribution.
Second, to encourage the prediction model to satisfy the MI robustness conditions, we propose alternative loss terms based on cross-entropy.
We theoretically show that minimizing each term ensures satisfaction of the corresponding component of the MI robustness conditions.
By combining all the loss terms, we formulate a new objective function, which we refer to as \textit{MIRRAMS (Mutual Information Regularization for Robustness Against Missingness Shifts)}.
An illustration of our method is provided in Figure \ref{fig:mirrams}.

The MIRRAMS offers appealing features. 
First, our method relies only on mild assumptions and is not restricted to specific missingness mechanisms such as MCAR, MAR, or MNAR, making it applicable to a wide range of real-world problems. 
Furthermore, it can readily handle cases where output labels are partially missing in the training data as well.
Under a weak condition on output missingness, we extend our method by adopting a semi-supervised learning (SSL) framework to train accurate models in such scenarios.


Interestingly, while deriving the new loss terms, we found that one of our proposed terms closely resembles the consistency regularization term originally introduced in FixMatch \citep{DBLP:conf/nips/SohnBCZZRCKL20}, one of the most widely used approaches for addressing SSL tasks.
As a by-product, our derivation thus provides a theoretical justification for the use of FixMatch and its subsequent variants.
	
We conduct extensive experiments on various benchmark datasets, including performance comparisons and ablation studies, and demonstrate that models trained with our method consistently outperform existing state-of-the-art frameworks for handling missing values across a variety of scenarios with missingness shifts between training and test-time.
Notably, MIRRAMS also achieves superior performance even when \textit{no missing} data are present, highlighting its effectiveness in standard supervised learning settings.
Furthermore, a simple modification of MIRRAMS extends its applicability to input-output-missing scenarios, where it continues to deliver strong and stable predictive performance, establishing MIRRAMS as a powerful framework for general-purpose tabular learning.

The remainder of this paper is organized as follows. 
We first review recent deep learning approaches relevant to our work.
Then, we introduce our proposed learning framework, MIRRAMS, and describe its core components and training objective in detail.
This is followed by extensive experimental analyses, including performance evaluations under various missingness scenarios and ablation studies.
Finally, we conclude with a summary of key findings and discuss potential directions for future research.
The main contributions of our work are:
\bed
    \item We propose mutual-information-based conditions, called \textit{MI robustness conditions}, to guide models to output robust predictions under missingness distribution shifts.
    \item We introduce a new, theoretically grounded learning framework, \textit{MIRRAMS}, designed to encourage models to satisfy the MI robustness conditions.
    \item We empirically demonstrate the superiority of MIRRAMS across various scenarios with missingness distribution shifts, using widely used benchmark datasets.
\eed

\section{Related Works}
\label{sec:related}
	
Despite their importance, missingness shifts have only recently gained attention, with earlier work mostly limited to illustrative or empirical studies \citep{groenwold2020informative}. 
\citet{ouyang2021maximum} introduced an MMD-based regularization to learn robust embeddings, and \citet{zhou2023domain} formally defined missingness shifts and related them to covariate shift tasks \citep{gretton2009covariate,DBLP:conf/icml/YuS12}.
NeuMISE \citep{rockenschaub2024robust} is an end-to-end model that embeds missingness patterns without imputation. 
However, all of these approaches are limited by their reliance on access to test-time inputs or distributional assumptions such as ignorability.

Now we focus on related work that deals with data containing missing values.
Roughly, existing deep learning approaches can be categorized into two groups: 1) imputing missing values and 2) treating missing values as masks.

Many methods adopt simple imputation, replacing missing continuous features with the mean and categorical ones with the mode or encoded tokens. Representative models include FT-Transformer \citep{DBLP:conf/nips/GorishniyRKB21}, SCARF \citep{bahri2021scarf}, ReConTab \citep{chen2023recontab}, XTab \citep{zhu2023xtab}, and SwitchTab \citep{wu2024switchtab}. TabPFN \citep{hollmann2022tabpfn} fills missing entries with zeros, which can be problematic when zero has semantic meaning. 
Alternatively, masking-based approaches treat missing values as learnable tokens. TabTransformer \citep{huang2020tabtransformer} uses a special embedding for missing categorical features, while SAINT \citep{DBLP:journals/corr/abs-2106-01342} adds binary missing indicators and dedicated embeddings. 

\textcolor{black}{
On the other hand, some recent models, such as VIME \citep{DBLP:conf/nips/YoonZJS20}, UniTabE \citep{yang2023unitabe}, and PMAE \citep{DBLP:conf/aaai/Kim0P25}, introduce random masking during training and adopt self-supervised or masked autoencoder-based objectives to improve robustness against missing data at test-time.
}

\section{Proposed Method}
\label{sec:method}
	
\subsection{Notations and Definitions}
\label{sec:pre}
	
We introduce notations and definitions used throughout the paper.
Let $(\boldsymbol{X}, Y) \in \mathcal{X} \times [K]$ be a pair of fully observed input and label random variables, where $\mathcal{X}$ represents the input space composed of $p$ variables, which may include both continuous and categorical features, and $[K] := {1, \ldots, K}$ denotes the set of label classes.

We denote the complete joint distribution of the input-output pairs by $\mathbb{P}$, which is assumed to be identical for both the training and test data.
The missingness operators for the training and test data are denoted by $M^{\text{tr}}$ and $M^{\text{ts}}$, respectively, and may depend on each data instance. 
Unless otherwise specified, $M^{\text{tr}}$ and $M^{\text{ts}}$ may be associated with input missingness.
We make no specific assumptions on $M^{\text{tr}}$ and $M^{\text{ts}}$, such as MCAR, MAR, or MNAR. 

The observed training dataset is given by $\mathcal{D}^{\text{tr}} := \{(\boldsymbol{x}_i,y_i),i=1,\ldots,n\}$, where each pair is drawn independently from $M^{\text{tr}} \circ \mathbb{P}$. 
Similarly, we assume that each input-output pair in the \textit{unseen} test dataset is independently drawn from the distribution $M^{\text{ts}} \circ \mathbb{P}$. 
For simplicity, we use the dataset notation interchangeably with its corresponding empirical distribution whenever there is no confusion.
In addition, when the context is clear, we use the one-hot encoded representation of the label, that is, $y\in\{0,1\}^K$, interchangeably with its scalar form.


We define an additional masking operator as $M_{\boldsymbol{r}}^{\text{m}}$, where $\boldsymbol{r} = (r_1, \ldots, r_p)^T \in (0,1)^p$ denotes a masking ratio vector for each variable.
Unless otherwise specified, we assume that $M_{\boldsymbol{r}}^{\text{m}}$ follows a product of $p$ independent Bernoulli distributions with a common ratio $r$, i.e., $M_{\boldsymbol{r}}^{\text{m}}=M_{r}^{\text{m}} = \text{Ber}(r) \otimes \cdots \otimes \text{Ber}(r)$, where a value of one indicates the variable is observed and zero indicates it is masked.

	
For a given input $\boldsymbol{x} = (x_1, \ldots, x_p)^T$, we define $v(\boldsymbol{x}) = (v_0([\texttt{CLS}]), v_1(x_1), \ldots, v_p(x_p))^T \in \mathbb{R}^{(p+1)\cdot d}$ as a concatenated embedding vector, where $[\texttt{CLS}]$ denotes the classification token and $v_0([\texttt{CLS}]) \in \mathbb{R}^d$ is its embedding vector.
Each $v_j(x_j) \in \mathbb{R}^d$, $j \in [p]$, represents the embedding function of the $j$-th input variable.
Following \citet{DBLP:journals/corr/abs-2106-01342}, we implement each $v_j$ using a multilayer perceptron (MLP) when $x_j$ is continuous, and a learnable token embedding when $x_j$ is categorical.
In addition, we introduce a separate learnable token embedding for each variable to explicitly represent its missingness.
	

Based on the concatenated embedding, a transformer-based encoder is denoted as ${enc}:\mathbb{R}^{(p+1)\cdot d} \to \mathbb{R}^{(p+1)\cdot d}$.
The encoder output corresponding to the classification token is used as the final representation of input $\boldsymbol{x}$ for prediction.
A prediction head $g:\mathbb{R}^d \to \mathcal{S}^K$ is applied to produce the final output, where $\mathcal{S}^K$ denotes the $K$-dimensional simplex.
The overall prediction model is thus defined as $f(\boldsymbol{x}) := g \circ enc \circ v(\boldsymbol{x})$.
The predicted label is given by $\hat{y}(\boldsymbol{x}) := \arg\max_{k \in [K]} f_k(\boldsymbol{x})$, where $f = (f_1, \ldots, f_K)^T$, and the confidence is $\hat{c}(\boldsymbol{x}) := \max_{k \in [K]} f_k(\boldsymbol{x})$.

For two random variables \( U \) and \( V \), we denote by \( H(U) \) the entropy, \( D_{\text{KL}}(U \| V) \) the Kullback-Leibler (KL) divergence, and \( H(U \mid V) \) the conditional entropy.
We also denote by \( \mathbb{I}(\cdot) \) the indicator function.



	
\subsection{Mutual-Information-Based Robustness Conditions} 

\paragraph{MI robustness}
Our goal is to train a prediction model $f$ that achieves high performance on \textit{unseen} test data corrupted with an \textit{unknown} missingness distribution.
To this end, we employ the mutual information measure and introduce the following conditions--termed \textit{MI robustness conditions}--formulated as:

\begin{align}
    \label{eq:infomin_ideal}
    &\begin{array}{c}
    \begin{array}{cc}
        \begin{array}{c}
            I\left(M^{\text{tr}}(\boldsymbol{X}); Y\right) \\[1ex]
            \vcenter{\hbox{\raisebox{0.4ex}{\scriptsize{$(i)$}}\hspace{0.2em}\rotatebox{90}{\(\approx\)}}} \\[1ex]       I\left(f(M^{\text{tr}}(\boldsymbol{X})); Y\right) \\[1.5ex]
            \vcenter{\hbox{\raisebox{-0.75ex}{\scriptsize{$(iii)$}}\hspace{0.2em}\rotatebox{-45}{\(\approx\)}}}
        \end{array}
        &
        \begin{array}{c}
            I\left(M^{\text{ts}}(\boldsymbol{X}); Y\right) \\[1ex]
            \vcenter{\hbox{\rotatebox{90}{\(\approx\)}\hspace{0.2em}\raisebox{0.4ex}{\scriptsize{$(ii)$}}}} \\[1ex]
            I\left(f(M^{\text{ts}}(\boldsymbol{X})); Y\right) \\[1.5ex]
            \vcenter{\hbox{\rotatebox{45}{\(\approx\)}\hspace{0.2em}\raisebox{0.4ex}{\scriptsize{$(iv)$}}}}
        \end{array}
    \end{array} \\[7.5ex]
    I\left(f(M^{\text{tr}}(\boldsymbol{X})); f(M^{\text{ts}}(\boldsymbol{X}))\right)
    \end{array}
\end{align}

The detailed explanation of the MI robustness conditions is as follows. 
The function $f$ should \textit{preserve} information regarding each input's label distribution under the missingness distribution both in the training and test datasets ((i) and (ii) in \eqref{eq:infomin_ideal}).
And the prediction function should \textit{discard} unnecessary information that is not relevant to predicting the label ((iii) and (iv) in \eqref{eq:infomin_ideal}).
Together, these conditions guide the model to extract only essential label-related information while remaining robust to various missingness patterns.
	

    \paragraph{Modified MI robustness}
However, the above conditions in \eqref{eq:infomin_ideal} are intractable to enforce directly, as the test-time data distribution, $M^{\text{ts}}\circ \mathbb{P}$, is inaccessible--unlike $M^{\text{tr}}$, for which training samples drawn from $M^{\text{tr}} \circ \mathbb{P}$ are available. 
To address this issue, we construct an \textit{enlarged} missing data distribution by applying the operator $M_{{r}}^{\text{m}}$ to the training data, resulting in $M_{{r}}^{\text{m}} \circ M^{\text{tr}}\circ \mathbb{P}$, and use it as a \textit{surrogate} for the test-time distribution.
By applying this approximation to \eqref{eq:infomin_ideal}, we derive the following \textit{modified} MI robustness conditions:
    \begin{align}
    \label{eq:infomin_oursetting}
        &\begin{array}{c}
        \begin{array}{cc}
            \begin{array}{c}
                I\left(M^{\text{tr}}(\boldsymbol{X}); Y\right) \\[1ex]
                \vcenter{\hbox{\raisebox{0.4ex}{\scriptsize{$(i)$}}\hspace{0.2em}\rotatebox{90}{\(\approx\)}}} \\[1ex]       I\left(f(M^{\text{tr}}(\boldsymbol{X})); Y\right) \\[1.5ex]
                \vcenter{\hbox{\raisebox{-0.75ex}{\scriptsize{$(iii)$}}\hspace{0.2em}\rotatebox{-45}{\(\approx\)}}}
            \end{array}
            &
            \begin{array}{c}
                I\left(M_{{r}}^{\text{m}}\circ M^{\text{tr}}(\boldsymbol{X}); Y\right) \\[1ex]
                \vcenter{\hbox{\rotatebox{90}{\(\approx\)}\hspace{0.2em}\raisebox{0.4ex}{\scriptsize{$(ii)$}}}} \\[1ex]
                I\left(f(M_{{r}}^{\text{m}}\circ M^{\text{tr}}(\boldsymbol{X})); Y\right) \\[1.5ex]
                \vcenter{\hbox{\rotatebox{45}{\(\approx\)}\hspace{0.2em}\raisebox{0.4ex}{\scriptsize{$(iv)$}}}}
            \end{array}
        \end{array} \\[7.5ex]
        I\left(f(M^{\text{tr}}(\boldsymbol{X})); f(M_{{r}}^{\text{m}}\circ M^{\text{tr}}(\boldsymbol{X}))\right)
        \end{array}
    \end{align}


Assuming that the surrogate data distribution sufficiently approximates that of the test data distribution, the prediction model \( f \) can be guided such that satisfying conditions (ii) and (iv) in \eqref{eq:infomin_oursetting} also implies the satisfaction of the corresponding conditions in \eqref{eq:infomin_ideal}. 
This, in turn, enforces the resulting model to be robust to distributional shifts in missingness.

We note that, as long as the operator \( M^{\text{tr}} \) is non-deterministic, the support of \( M_{r}^{\text{m}} \circ M^{\text{tr}} \circ \mathbb{P} \) always covers that of \( M^{\text{ts}} \circ \mathbb{P} \) for any \( r > 0 \), validating the use of \( M_{r}^{\text{m}} \) to construct a surrogate distribution. 
However, a poor choice of $r$ may cause a notable discrepancy between the simulated and actual test-time missingness distributions, weakening the rationale for \eqref{eq:infomin_oursetting} and harming performance.
Therefore, selecting an appropriate value of \( r \) is critical for the effectiveness of the proposed MI robustness conditions. 
In practice, however, setting \( r \) to any reasonable positive value is generally sufficient as long as it is neither too small nor too large, which we empirically validate in the ablation studies.
We emphasize again that no specific assumptions on the missingness mechanism, such as MCAR, MAR, or MNAR, are required to establish the conditions in \eqref{eq:infomin_oursetting}.

\begin{remark}
The reference \citep{DBLP:conf/nips/Tian0PKSI20} introduced the \textit{InfoMin} principle, which uses mutual information as a criterion for designing encoders that yield optimal representations for downstream prediction tasks.
Since computing mutual information exactly is intractable, they approximated it using the InfoNCE loss \citep{oord2018representation}, a widely used lower bound in contrastive learning.
In contrast, we propose a different and effective strategy to satisfy mutual information-based conditions without relying on contrastive learning, enabling efficient model training under our setting.
\end{remark}

\subsection{Derivation of Loss Function in MIRRAMS}
\label{sec:loss}
Now we propose new loss functions to encourage the prediction model $f$ to satisfy the modified {MI robustness conditions} in \eqref{eq:infomin_oursetting}.  
Assuming that $I\left(M^{\text{tr}}(\boldsymbol{X}); Y\right)\approx I\left(M_{{r}}^{\text{m}}\circ M^{\text{tr}}(\boldsymbol{X}); Y\right)$ with a carefully chosen masking ratio $r$, it is sufficient to introduce three loss terms, each corresponding to conditions (i), (ii), and (iii) in \eqref{eq:infomin_oursetting}. 
	
\paragraph{Conditions (i) \& (ii)}
For the condition (i), since even approximating $I\left(M^{\text{tr}}(\boldsymbol{X}); {Y}\right)$ is intractable, directly enforcing similarity between the two mutual information values is not feasible. 
To address this, we use the \textit{conditional entropy} between $Y$ and $f(M^{\text{tr}}(\boldsymbol{X}))$ as a surrogate loss term, given as:
{\fontsize{9.5pt}{11pt}\selectfont
\begin{align}
\label{eq:ce}
  H(Y \mid f(M^{\text{tr}}(\boldsymbol{X}))) :=
  {\mathbb{E}}_{M^{\text{tr}}(\boldsymbol{X}),Y}
  \left[-\log P\left(Y \mid f(M^{\text{tr}}(\boldsymbol{X}))\right)\right]
\end{align}
}
The following proposition provides justification for using the conditional entropy to satisfy the condition (i). 
The proof is deferred to the Appendix A.

\begin{proposition}
    \label{prop1}
    Consider a pair of two random variables $(U,V)$ and a function $\xi$. 
    Suppose there exists a positive constant $\epsilon>0$ such that 
    \begin{align*}
        D_{KL}\left(P(V|U=u)||P(V|\xi(U)=\xi(u))\right)<\epsilon    
    \end{align*}
    for all $u\in\text{supp}(U)$. 
    Then, the following holds:
    \begin{align*}
        \left| I(U;V)-I(\xi(U);V)\right|<\epsilon.
    \end{align*}
\end{proposition}

We next present the implication of Proposition \ref{prop1}. 
The conditional entropy in \eqref{eq:ce} can be reformulated as:
\begin{align*}
&H(Y|f(M^{\text{tr}}(\boldsymbol{X}))) = H(Y|M^{\text{tr}}(\boldsymbol{X}))+\\
&\quad{\mathbb{E}}_{M^{\text{tr}}(\boldsymbol{X})}\left[D_{KL}\left(P(Y|M^{\text{tr}}(\boldsymbol{X}))||P(Y|f(M^{\text{tr}}(\boldsymbol{X})))\right)\right],
\end{align*}
where the first term does not depend on the model $f$. 
Thus, minimizing \eqref{eq:ce} with respect to $f$ corresponds to finding a function $f$ such that the conditional distribution $P(Y|f(\tilde{\boldsymbol{x}}))$ closely approximates $P(Y|\tilde{\boldsymbol{x}})$ in terms of KL divergence, for all $\tilde{\boldsymbol{x}}\sim M^{\text{tr}}(\boldsymbol{X})$. 
By Proposition \ref{prop1}, such a function $f$ satisfies condition (i) in \eqref{eq:infomin_oursetting}, thereby validating the use of \eqref{eq:ce} as a surrogate loss for it.
In practice, \eqref{eq:ce} is approximated by the empirical average of the cross-entropy computed over the training dataset, given by:
\begin{align}
    \label{eq:ce_tr}
    \mathcal{L}_1^{\text{MI}} := 
    \mathbb{E}_{(\tilde{\boldsymbol{X}},Y)\sim\mathcal{D}^{\text{tr}}}\bigl[-\log P(Y|f(\tilde{\boldsymbol{X}}))\bigr].
\end{align}
	
The fulfillment of the condition (ii) can be achieved similarly to condition (i), by minimizing the conditional entropy $H(Y|f(M_{{r}}^{\text{m}}\circ M^{\text{tr}}(\boldsymbol{X})))$. 
This quantity can likewise be approximated by the empirical average of the cross-entropy, formulated as:
\begin{align}
    \label{eq:ce2_tr}
    \mathcal{L}_2^{\text{MI}} := \mathbb{E}_{(\tilde{\boldsymbol{X}},Y)\sim\mathcal{D}^{\text{tr}}}
    {\mathbb{E}}_{M_{{r}}^{\text{m}}}\bigl[-\log P(Y|f(M_{{r}}^{\text{m}}(\tilde{\boldsymbol{X}})))\bigr].
\end{align}
	
\paragraph{Condition (iii)}
Lastly, we focus on the condition (iii) in \eqref{eq:infomin_oursetting}. 
We note that the right-hand side of the condition (iii) can be decomposed as:
\begin{align}
    \label{eq:rhs_iii}
    &I\left(f(M^{\text{tr}}(\boldsymbol{X})); f(M_{{r}}^{\text{m}}\circ M^{\text{tr}}(\boldsymbol{X}))\right)\nonumber\\
    &=H(f(M^{\text{tr}}(\boldsymbol{X})))-H\left(f(M^{\text{tr}}(\boldsymbol{X}))| f(M_{{r}}^{\text{m}}\circ M^{\text{tr}}(\boldsymbol{X}))\right).
\end{align}
Suppose that the training procedure has \textit{sufficiently progressed} such that $f(M^{\text{tr}}({\boldsymbol{x}}))$ captures nearly all the information relevant to its corresponding label, $y$. 
That implies $f(M^{\text{tr}}(\boldsymbol{\boldsymbol{x}}))$ becomes nearly a one-hot encoded $y$, and therefore the information contained in $f(M^{\text{tr}}(\boldsymbol{\boldsymbol{X}}))$ and $\hat{y}(M^{\text{tr}}(\boldsymbol{\boldsymbol{X}}))$ becomes \textit{almost equivalent}. 
As a result, the mutual information in \eqref{eq:rhs_iii} can be approximated by 
\begin{align*}
    &I\left(f(M^{\text{tr}}(\boldsymbol{X})); f(M_{{r}}^{\text{m}}\circ M^{\text{tr}}(\boldsymbol{X}))\right)\\
    & \quad\approx H(\hat{y}(M^{\text{tr}}(\boldsymbol{X})))-H\left(\hat{y}(M^{\text{tr}}(\boldsymbol{X}))| f(M_{{r}}^{\text{m}}\circ M^{\text{tr}}(\boldsymbol{X}))\right)\nonumber\\
    &\quad\approx H(Y)-H\left(\hat{y}(M^{\text{tr}}(\boldsymbol{X}))| f(M_{{r}}^{\text{m}}\circ M^{\text{tr}}(\boldsymbol{X}))\right).
\end{align*}
	
	
Therefore, using the fact that $I\left(f(M^{\text{tr}}(\boldsymbol{X})); Y\right)=H(Y)-H(Y|f(M^{\text{tr}}(\boldsymbol{X})))$, condition (iii) in \eqref{eq:infomin_oursetting} reduces to the following: 
\begin{align}
    \label{eq:rhs_iii_app}
    H(Y|f(M^{\text{tr}}(\boldsymbol{X})))\approx
    H\left(\hat{y}(M^{\text{tr}}(\boldsymbol{X}))| f(M_{{r}}^{\text{m}}\circ M^{\text{tr}}(\boldsymbol{X}))\right).
\end{align}
Since encouraging condition (i) leads to minimization of  $H(Y|f(M^{\text{tr}}(\boldsymbol{X})))$, the condition in \eqref{eq:rhs_iii_app} is satisfied when $H\left(\hat{y}(M^{\text{tr}}(\boldsymbol{X}))| f(M_{{r}}^{\text{m}}\circ M^{\text{tr}}(\boldsymbol{X}))\right)$ is also minimized. 
Accordingly, we define the following loss term:
\begin{align}
    \label{eq:ce3_tr}
    &\mathcal{L}_3^{\text{MI}} := 
    \mathbb{E}_{\tilde{\boldsymbol{X}}\sim\mathcal{D}_x^{\text{tr}}}
    {\mathbb{E}}_{M_{{r}}^{\text{m}}}
    \bigl[-\log P(\hat{y}(\tilde{\boldsymbol{X}})|f(M_{{r}}^{\text{m}}(\tilde{\boldsymbol{X}}))) \nonumber\\
    &\quad\quad\quad\quad\quad\quad\quad\quad\quad\times \mathbb{I}(\hat{c}(\tilde{\boldsymbol{X}})\ge \tau)\bigr],
\end{align}
where $\tau\in(0,1)$ is a predifined confidence threshold and $\mathcal{D}_x^{\text{tr}}$ is the training input set.
Minimizing \eqref{eq:ce3_tr} promotes the reduction of $H\left(\hat{y}(M^{\text{tr}}(\boldsymbol{X}))| f(M_{{r}}^{\text{m}}\circ M^{\text{tr}}(\boldsymbol{X}))\right)$, but this holds only when the overall predictions are made with \textit{high confidence}.
Such confidence typically emerges when the model $f$ is \textit{sufficiently trained}, which is consistent with the the assumptions under which condition in \eqref{eq:rhs_iii_app} is derived.
	
	
\paragraph{Final Loss Function}
By combining the loss terms defined in \eqref{eq:ce_tr}, \eqref{eq:ce2_tr}, and \eqref{eq:ce3_tr}, the final objective function of our proposed method is formulated as:
\begin{align}
    \label{eq:final_loss1}
    \mathcal{L}:=\mathcal{L}_1^{\text{MI}}+\lambda_1\cdot \mathcal{L}_2^{\text{MI}} + \lambda_2\cdot \mathcal{L}_3^{\text{MI}},
\end{align}
where $\lambda_1,\lambda_2>0$ are hyperparameters. 
We estimate the parameters of the prediction model $f$ by minimizing the loss $\mathcal{L}$ using a gradient-descent-based optimizer, such as Adam \citep{kingma2014adam}. 
In practice, the optimal combination of hyperparameters is selected using a validation dataset. 
	

\subsection{Extension to Missing Labels Scenario}
In the previous sections, we have considered scenarios where missingness occurs only in the input features.  
In this section, we show that our method can be naturally extended to settings where \textit{output labels} are also partially missing in training data.
We assume that the output is missing under MCAR or MAR mechanisms, meaning that the output missingness is not dependent on the output itself.
Under such conditions, it is well established that training prediction models reduces to a {semi-supervised learning} (SSL) problem without requiring bias correction \citep{chapelle2006}.
Accordingly, we extend our method to the conventional SSL setting to handle this case.

We denote by \( \mathcal{D}_l^{\text{tr}}\) and \(\mathcal{D}_u^{\text{tr}}\) the subsets of the training data with observed and missing outputs, respectively.  
In the loss function $\mathcal{L}$ defined in \eqref{eq:final_loss1}, the first two terms, \( \mathcal{L}_1^{\text{MI}} \) and \( \mathcal{L}_2^{\text{MI}} \), are computed using both inputs and outputs, while the third term, \( \mathcal{L}_3^{\text{MI}} \), is computed solely based on the inputs.  
Therefore, our approach can be naturally extended to output-missing scenarios by computing the expectations of \( \mathcal{L}_1^{\text{MI}} \) and \( \mathcal{L}_2^{\text{MI}} \) over \( \mathcal{D}_l^{\text{tr}} \), and that of \( \mathcal{L}_3^{\text{MI}} \) over \( \mathcal{D}_u^{\text{tr}} \).
In the experimental section, we demonstrate that this modified version of MIRRAMS maintains strong performance, even when the majority of training labels are missing.


\subsection{Rethinking Consistency Regularization-Based SSL}

As a by-product of the loss function derivation of MIRRAMS, we provide a theoretical perspective on SSL methods that employ \textit{consistency regularization}. 
{Consistency regularization} is a core principle in SSL domain, which encourages a prediction model to produce \textit{similar} predictions for the same input under \textit{different} augmentations.
One of the most prominent approaches built on this principle is FixMatch \citep{DBLP:conf/nips/SohnBCZZRCKL20}, whose objective function is:
$$\mathcal{L}^{\text{CE}}+\lambda\cdot\mathcal{L}^{\text{FM}},$$
where 
\begin{align*}
    \mathcal{L}^{\text{CE}} := \mathbb{E}_{(\boldsymbol{X},Y)\sim\mathcal{L}^{\text{tr}}}
    {\mathbb{E}}_{\alpha}\left[-\log P\left(Y|f(\alpha(\boldsymbol{X}))\right)\right],  
\end{align*}
\begin{align*}
    &\mathcal{L}^{\text{FM}} := \mathbb{E}_{\boldsymbol{X}\sim\mathcal{U}^{\text{tr}}}
    {\mathbb{E}}_{\alpha,\mathcal{A}}\bigl[-\log P\left(\hat{y}(\alpha(\boldsymbol{X}))|f(\mathcal{A}(\boldsymbol{X}))\right)\\
    &\quad\quad\quad\quad\quad\quad\quad\quad\quad\times \mathbb{I}(\hat{c}(\alpha(\boldsymbol{X}))\ge \tau)\bigr],
\end{align*}
$\mathcal{L}^{\text{tr}}$ and $\mathcal{U}^{\text{tr}}$ denote the empirical distributions of the labeled and unlabeled training data, respectively, and $\lambda>0, \tau\in(0,1)$ are two hyperparameters.
Here, $\alpha$ and $\mathcal{A}$ represent weak and strong augmentations, respectively. 
The authors of FixMatch intuitively interpret minimizing this term as encouraging the model $f$ to correctly classify labeled data, while enforcing consistency between predictions under weak and strong augmentations, conditioned on high-confidence predictions from the weakly augmented inputs.
	
On the other hand, from our perspective, this loss function can be interpreted as guiding the prediction model $f$ to satisfy the following mutual information conditions:
{\fontsize{9.5pt}{11pt}\selectfont
\begin{align}
    \label{eq:mi_conds_fm}
    I(Y;\alpha(\boldsymbol{X}))\underset{(i)}{\approx} I(Y;f(\alpha(\boldsymbol{X})))
    \underset{(ii)}{\approx} I(f(\alpha(\boldsymbol{X}));f(\mathcal{A}(\boldsymbol{X}))).
\end{align}
}
Condition (i) in \eqref{eq:mi_conds_fm} encourages a prediction model $f$ to extract label-relevant information, while condition (ii) prevents it from encoding redundant information unrelated to label prediction, regardless of the augmentation type.
According to our theoretical analysis, minimizing $\mathcal{L}^{\text{CE}}$ and $\mathcal{L}^{\text{FM}}$ promotes the satisfaction of conditions (i) and (ii) in \eqref{eq:mi_conds_fm}, respectively.
Therefore, FixMatch can be viewed as training a prediction model to extract label-relevant information while suppressing redundant features, even under strong augmentations.
This alternative interpretation provides a more theoretically grounded explanation for the empirical success of FixMatch, as well as for subsequent SSL methods inspired by it 
\citep{DBLP:conf/icml/XuSYQLSLJ21,DBLP:conf/nips/ZhangWHWWOS21,DBLP:conf/icml/GuoL22,DBLP:conf/iclr/Wang0HHFW0SSRS023}.

\section{Experiments}
\label{sec:exp}

We validate the effectiveness of our method under distributional shifts in missingness patterns between training and test datasets.
Additionally, we demonstrate that our method achieves superior performance even in standard supervised settings where no missing values are present.
We further demonstrate that, with a simple modification, our method remains highly effective when outputs are also missing, highlighting its adaptability.
All results are averaged over three trials with different random initializations.
Our implementation is based on the \texttt{PyTorch} framework and runs on two NVIDIA GeForce RTX 3090 GPUs.

\paragraph{Dataset Description}
We conduct experiments on ten widely used benchmark datasets, all of which are publicly available from either the UCI repository or AutoML benchmarks.
Each dataset is split into training, validation, and test sets with a ratio pair of $(0.65, 0.15, 0.2)$. 
Detailed information about the datasets is provided in the Appendix B.

To evaluate robustness under diverse missingness scenarios, we consider three settings: (1) missing completely at random (MCAR), (2) missing at random (MAR), and (3) missing not at random (MNAR).
In the MCAR setting, each input variable is independently missing with equal probability.
For the MAR setting, input variables are divided into two partitions: the first partition is missing according to the MCAR mechanism, while variables in the second partition are missing with probabilities computed based on the observed values from the first partition.
In the MNAR setting, we follow the same procedure as in the MAR case, except that the missing probabilities for the second partition are computed using both the observed and missing variables.

The marginal missing probability for each variable is set to $\alpha$.  
We vary $\alpha$ across both training and test datasets, specifically setting $\alpha^{\text{tr}},\alpha^{\text{ts}} \in \{0, 0.15, 0.3\}$.
Note that the setting $(\alpha^{\text{tr}},\alpha^{\text{ts}})=(0,0)$ corresponds to the standard supervised learning setting, where neither the training nor the test data contain any missing values. 
The validation dataset shares the same missingness pattern as the training dataset.
Details of the missingness generation are provided in Appendix B.
    

    \begin{table*}[t]
    \centering
    \caption{Comparison of averaged test AUC scores (\%) across 10 tabular datasets under various combinations of $(\alpha^{\text{tr}},\alpha^{\text{ts}})$. 
    All results were obtained using our own implementations. 
    For each setting, the best result is highlighted in {bold}, and the second-best is {underlined}.}
    \resizebox{0.99\textwidth}{!}{%
    \begin{tabular}{l|c|cccccccc|cccccccc}
    \toprule
    \textbf{Method} &
    \textbf{No Missing} &
    \multicolumn{8}{c|}{\textbf{MAR}} &
    \multicolumn{8}{c}{\textbf{MNAR}} \\ \midrule
    \textbf{$\alpha^{\text{tr}}$} &
    0 &
    \multicolumn{2}{c|}{0} &
    \multicolumn{3}{c|}{0.15} &
    \multicolumn{3}{c|}{0.3} &
    \multicolumn{2}{c|}{0} &
    \multicolumn{3}{c|}{0.15} &
    \multicolumn{3}{c}{0.3} \\ \midrule
    \textbf{$\alpha^{\text{ts}}$} &
    0 &
    0.15 &
    \multicolumn{1}{c|}{0.3} &
    0 &
    0.15 &
    \multicolumn{1}{c|}{0.3} &
    0 &
    0.15 &
    0.3 &
    0.15 &
    \multicolumn{1}{c|}{0.3} &
    0 &
    0.15 &
    \multicolumn{1}{c|}{0.3} &
    0 &
    0.15 &
    0.3 \\ \midrule
    LR-0 &
      72.03 &
      67.35 &
      \multicolumn{1}{c|}{63.85} &
      70.31 &
      67.94 &
      \multicolumn{1}{c|}{65.32} &
      69.09 &
      67.57 &
      65.12 &
      67.30 &
      \multicolumn{1}{c|}{64.26} &
      69.64 &
      67.28 &
      \multicolumn{1}{c|}{65.95} &
      68.06 &
      66.51 &
      65.06 \\
    NN-0 &
      77.41 &
      74.24 &
      \multicolumn{1}{c|}{69.37} &
      80.10 &
      78.58 &
      \multicolumn{1}{c|}{77.10} &
      79.84 &
      78.74 &
      77.98 &
      73.80 &
      \multicolumn{1}{c|}{69.75} &
      80.27 &
      78.46 &
      \multicolumn{1}{c|}{76.52} &
      79.10 &
      78.07 &
      77.10 \\
    RF &
      88.19 &
      83.96 &
      \multicolumn{1}{c|}{81.59} &
      85.87 &
      85.03 &
      \multicolumn{1}{c|}{82.94} &
      84.81 &
      84.18 &
      80.95 &
      84.45 &
      \multicolumn{1}{c|}{80.96} &
      86.08 &
      85.20 &
      \multicolumn{1}{c|}{83.67} &
      84.98 &
      83.61 &
      82.17 \\
    XGB &
      \underline{91.17} &
      82.33 &
      \multicolumn{1}{c|}{75.33} &
      \textbf{90.31} &
      \underline{88.58} &
      \multicolumn{1}{c|}{\underline{86.29}} &
      \underline{89.65} &
      \underline{88.10} &
      \underline{86.16} &
      82.82 &
      \multicolumn{1}{c|}{75.22} &
      \underline{89.71} &
      88.09 &
      \multicolumn{1}{c|}{85.57} &
      88.89 &
      87.17 &
      85.57 \\
    CB &
      89.84 &
      \underline{87.19} &
      \multicolumn{1}{c|}{\underline{83.96}} &
      88.89 &
      86.89 &
      \multicolumn{1}{c|}{84.32} &
      88.68 &
      87.02 &
      85.62 &
      \underline{86.84} &
      \multicolumn{1}{c|}{\underline{83.00}} &
      \underline{89.71} &
      \underline{88.54} &
      \multicolumn{1}{c|}{\underline{86.83}} &
      \underline{88.97} &
      \underline{87.54} &
      \underline{85.83} \\ \midrule
    SAINT &
      90.03 &
      85.44 &
      \multicolumn{1}{c|}{82.35} &
      89.16 &
      88.01 &
      \multicolumn{1}{c|}{85.89} &
      88.66 &
      87.63 &
      85.86 &
      84.86 &
      \multicolumn{1}{c|}{81.76} &
      89.07 &
      87.31 &
      \multicolumn{1}{c|}{85.51} &
      88.40 &
      86.89 &
      85.26 \\
    TabTF &
      86.83 &
      83.35 &
      \multicolumn{1}{c|}{81.66} &
      84.49 &
      82.82 &
      \multicolumn{1}{c|}{80.95} &
      83.43 &
      81.98 &
      80.38 &
      83.18 &
      \multicolumn{1}{c|}{81.33} &
      84.74 &
      82.61 &
      \multicolumn{1}{c|}{80.55} &
      84.22 &
      81.86 &
      79.80 \\
    Stab &
      85.40 &
      82.44 &
      \multicolumn{1}{c|}{79.35} &
      85.37 &
      82.56 &
      \multicolumn{1}{c|}{80.24} &
      84.16 &
      80.89 &
      79.16 &
      82.85 &
      \multicolumn{1}{c|}{78.40} &
      85.81 &
      82.89 &
      \multicolumn{1}{c|}{79.19} &
      84.38 &
      81.53 &
      79.45 \\ \midrule
    Ours &
      \textbf{91.88} &
      \textbf{89.26} &
      \multicolumn{1}{c|}{\textbf{87.58}} &
      \underline{90.14} &
      \textbf{88.99} &
      \multicolumn{1}{c|}{\textbf{87.50}} &
      \textbf{89.75} &
      \textbf{88.53} &
      \textbf{87.08} &
      \textbf{89.20} &
      \multicolumn{1}{c|}{\textbf{87.86}} &
      \textbf{89.94} &
      \textbf{88.81} &
      \multicolumn{1}{c|}{\textbf{87.67}} &
      \textbf{89.60} &
      \textbf{88.50} &
      \textbf{87.35} \\ \bottomrule
    \end{tabular}}
    \label{tab:sl_result}
    \end{table*}

\begin{figure*}
    \centering
    \captionsetup{font=small}
    \includegraphics[width=0.95\linewidth]{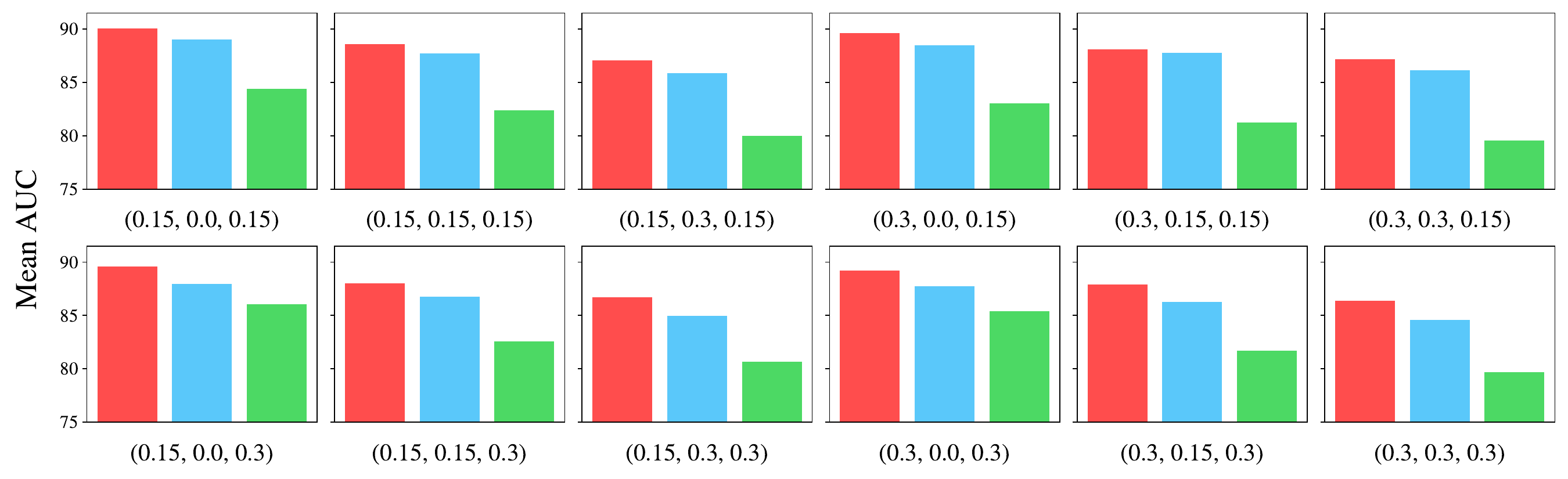}
    \caption{Comparison of averaged test AUC scores (\%) across 10 tabular datasets under settings with MCAR output missingness and MAR input missingness. 
    We compare three methods: ({\bf red}) Ours, ({\bf blue}) SAINT, and ({\bf green}) SwitchTab. In each panel, the x-axis denotes the combination of $(\alpha^{\text{tr}},\alpha^{\text{ts}},\alpha_y^{\text{tr}})$.}
    \label{fig:ssl_output}
\end{figure*}

\paragraph{Implementation Details}
We follow a similar architecture to that proposed by \citep{DBLP:journals/corr/abs-2106-01342}.
Specifically, for the embedding layers, we use an MLP for each continuous variable and learnable tokenized vectors for each categorical variable. 
And we also adopt the transformer architecture \citep{vaswani2017attention} for the encoder and a separate MLP for the prediction head. 
Detailed descriptions of the architectures we employ are stated in Appendix B.

As for the hyperparameters, we use the fixed configuration of $(r, \lambda_1, \lambda_2, \tau)=(0.2, 15, 15, 0.95)$ throughout all experiments. 
For training, we employ the Adam optimizer \citep{kingma2014adam} with a learning rate of $10^{-4}$, a mini-batch size of 256, and up to 1,000 epochs. 
The best model is selected based on performance on the validation dataset.

	

\subsection{Performance Results}
\label{sec:sl_case}
This section demonstrates that our method trains prediction models that are robust to various cases of missingness distributional shifts.
For comparison, we first consider five classical machine learning methods: (1) logistic regression with zero-value imputation (LR-0), (2) vanilla MLP with zero imputation (NN-0), (3) Random Forest (RF; \citet{breiman2001random}), (4) XGBoost (XGB; \citet{chen2016xgboost}), and (5) CatBoost (CB; \citet{DBLP:conf/nips/ProkhorenkovaGV18}). 
We implement these baselines using official \texttt{Python} packages, including \texttt{xgboost} and \texttt{scikit-learn}.
And we also consider three recent deep learning-based methods, all of which provide publicly available \texttt{GitHub} repositories: (6) SAINT \citep{DBLP:journals/corr/abs-2106-01342}, (7) TabTransformer (TabTF; \citet{huang2020tabtransformer}), and (8) SwitchTab (STab; \citet{wu2024switchtab}). 


    \begin{figure*}[t]
        \centering
        \captionsetup{font=small}
        \includegraphics[width=0.98\linewidth]{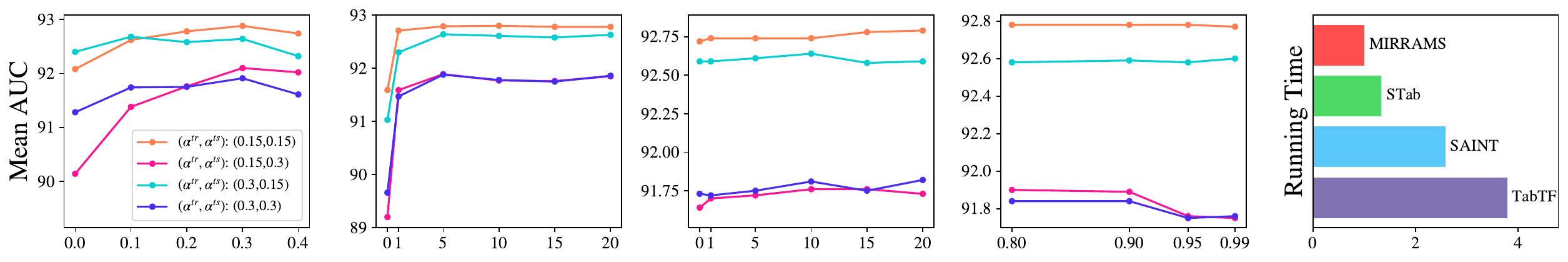}
        \caption{(\textbf{1st to 4th}) Test AUC scores for varying values of $r$, $\lambda_1$, $\lambda_2$, and $\tau$, respectively.
        (\textbf{5th}) Running time comparison. Each runtime is rescaled relative to that of MIRRAMS.}
        \label{fig:ablation}
    \end{figure*}
    
\paragraph{Case 1: Input Missingness}
For each setting, we evaluate the averaged test AUC score of our proposed method across ten datasets and compare it against the baseline methods.
The average results for the MAR and MNAR settings are summarized in Table \ref{tab:sl_result}, and detailed per-dataset results, including standard deviations and results for the MCAR, are provided in Appendix B. 

We observe that our method consistently outperforms all baseline methods, regardless of the type of missingness shift pattern.
Moreover, it yields stable AUC performance even as the missingness ratio in the test dataset increases, whereas other methods relatively exhibit a sharp performance drop.
These results highlight that our method enables the prediction model to extract label-relevant information while suppressing the influence of missingness patterns, thereby enhancing robustness to unseen missingness shifts.

Interestingly, our method also achieves improved performance in the standard scenario where neither the training nor test data contain missing values, i.e., $(\alpha^{\text{tr}},\alpha^{\text{ts}})=(0,0)$.
This observation can be attributed to the fact that applying additional masking during training acts as an \textit{implicit regularizer}, preventing the model from extracting excessive information that may lead to overfitting.
This conjecture is empirically supported: setting either $\lambda_1$ or $\lambda_2$ to zero results in lower average AUC scores of 90.81 and 91.35, respectively, which are inferior to the score of 91.88 achieved when both are non-zero.
These findings suggest that the two loss terms introduced by additional masking contribute to improved generalization by acting as an implicit regularizer.

As a result, our approach is not only effective under missingness shifts, but also performs well in standard supervised settings.
Furthermore, while other deep learning methods employ auxiliary decoder networks to boost performance, our method remains both effective and computationally efficient, requiring only a prediction model.

\paragraph{Case 2: Input-Output Missingness}

We focus on the MCAR output-missing case, as we observe that the results under the MAR condition exhibit similar overall trends.
Figure \ref{fig:ssl_output} presents the average AUC scores of our method under MAR input-missing scenarios with MCAR output-missing, across varying $(\alpha^{\text{tr}}, \alpha^{\text{ts}},\alpha_y^{\text{tr}})$ combinations, where $\alpha_y^{\text{tr}}$ denotes the training output missingness ratio.
The results are compared against SAINT and SwitchTab, two state-of-the-art SSL deep learning-based methods for tabular data.
Exact AUC values and results for the MCAR and MNAR input-missing scenarios are provided in Appendix B.
    
Our method consistently performs better than the baselines across all scenarios.
In addition, it relatively maintains stable decrease of AUC scores as the missingness ratio increases, while the performance of other methods degrades sharply. 
This further demonstrates the robustness of our approach under input-output-missing settings.
Combined with the results from input-missingness scenarios, these findings underscore the versatility of our method, positioning it as an off-the-shelf solution for a wide range of applications.


\subsection{Ablation Studies}
To examine the sensitivity of MIRRAMS to hyperparameter choices, we run additional experiments on three datasets--\texttt{adult}, \texttt{htru2}, and \texttt{qsar\_bio}--and report the average test AUC scores across four MNAR settings, i.e., $\alpha^{\text{tr}},\alpha^{\text{ts}}\in\{0.15, 0.3\}$.
For each dataset, we vary a single hyperparameter while keeping the others fixed at their optimal values.
The results is presented in Figure \ref{fig:ablation}, with further details provided in Appendix C. 
To summarize: 
\begin{itemize}
\item A value of $r$ greater than zero generally leads to improved performance, but performance deteriorates when $r$ becomes too large.
\item Regarding $\lambda_1$ and $\lambda_2$, our method performs well when both are set to positive values, indicating that the loss terms $\mathcal{L}_2^{\text{MI}}$ and $\mathcal{L}_3^{\text{MI}}$ contribute to performance improvement. 
However, their impacts exhibit slightly different trends: increasing $\lambda_1$ up to 5 results in steady performance improvements before reaching saturation, whereas increasing $\lambda_2$ also improves performance, but with relatively smaller gains.
\item Performance remains similar across all considered values of $\tau$, indicating that our method is not sensitive to its choice as long as a reasonable value is used.
\item  MIRRAMS maintains faster running time compared to other deep learning–based methods, in a case exceeding a three times speedup. 
This efficiency is attributed to the simplicity of the loss computation and the absence of auxiliary architectures such as decoders.
\end{itemize}

\section{Concluding Remarks}
\label{sec:conclusion}

We proposed MIRRAMS, a deep learning framework for robust prediction under distributional shifts in missingness patterns.
By introducing mutual information-based robustness conditions and corresponding loss terms, MIRRAMS guides models to extract label-relevant information that is robust to unseen missingness patterns.
As a by-product, our method provides a theoretical justification for the effectiveness of consistency regularization-based SSL methods, such as FixMatch.
Extensive experiments on ten benchmark datasets demonstrated that MIRRAMS consistently outperforms recent methods designed to handle missing values--across various missingness shift scenarios, and even when no missing data are present in either the training or test set.
Furthermore, we showed that MIRRAMS can be readily extended to scenarios where output labels are partially missing.
	
We note that our method does not employ any additional architectures, such as decoders, during training to improve prediction performance, whereas many recent approaches adopt \citep{chen2023recontab,wu2024switchtab}.
Therefore, combining our framework with auxiliary architectures used in other methods could further enhance performance, which is a promising direction for future research. 
Another interesting avenue is to investigate more refined masking strategies by searching for feature-wise masking ratios $\boldsymbol{r} = (r_1, \ldots, r_p)^T$ in $M_{\boldsymbol{r}}^{\text{m}}$, instead of relying on a single scalar ratio $r$, to achieve better coverage of test-time missingness patterns.
Furthermore, extending our approach to handle output-missingness under MNAR conditions by developing a new bias-correction framework would also be an intriguing avenue for future work.
\bibliography{references}

\appendix

\onecolumn

\vskip 0.2in
\begin{center}
    {\huge \textbf{Supplementary material for}}
    \\
    {\huge \textbf{``MIRRAMS: Learning Robust Tabular Models under Unseen Missingness Shifts''}}
\end{center}
\vskip 0.2in

\section{A. Proof of Proposition 1}
\label{sec:appA}

It is sufficient to prove the result for the case where both $U$ and $V$ are discrete random variables. 
The extension to the continuous case follows straightforwardly. 
The mutual information can be reformulated as:
\begin{align*}
	I(U;V)=H(V)-H(V|U)\quad\text{and}\quad I(\xi(U);V)=H(V)-H(V|\xi(U)),
\end{align*}
where $H(\cdot|\cdot)$ denotes the conditional entropy. 
By the definition of the conditional entropy, we have
\begin{align}
	\label{eq:app1}
	H(V|U)=\sum_{u,v}p(u,v)\cdot\log\frac{p(u)}{p(u,v)}=\sum_{u,v}p(u,v)\cdot\left[-\log p(v|u)\right],
\end{align}
and similarly,
\begin{align}
	\label{eq:app2}
	H(V|\xi(U))=\sum_{u,v}p(u,v)\cdot\log\frac{p(\xi(u))}{p(\xi(u),v)}=\sum_{u,v}p(u,v)\cdot\left[-\log p(v|\xi(u))\right].
\end{align}
By subtracting \eqref{eq:app1} from \eqref{eq:app2}, we obtain
\begin{align}
	I(U;V)-I(\xi(U);V)&=H(V|\xi(U))-H(V|U) \nonumber \\
	&=\sum_{u,v}p(u,v)\cdot\log\left[ \frac{p(v|u)}{p(v|\xi(u))}\right] \nonumber \\
	&=\sum_{u}p(u)\sum_{v}p(v|u)\cdot\log\left[ \frac{p(v|u)}{p(v|\xi(u))}\right] \nonumber\\
	&=\mathbb{E}_U\left[ D_{KL}\left(P(V|U=u)||P(V|\xi(U)=\xi(u))\right) \right]\nonumber\\
	&<\epsilon. \nonumber
\end{align}
By the data processing inequality \citep{beaudry2011intuitive}, it is known that $I(U;V)\ge I(\xi(U);V)$ for any deterministic function $\xi$. 
Hence, the proof is complete. \qed

\newpage
\section{B. Benchmark Dataset Information and Experiment Results}
\label{sup:experiment_results}

\subsection{B-\MakeUppercase{\romannumeral 1}. Benchmark Dataset Information}
\label{sec:appB1}

We evaluate a total of 10 tabular datasets. 
All datasets are publicly available and can be obtained from sources such as UCI, AutoML, or Kaggle competitions, as listed in Table \ref{sup:tab:dataset_urls}. 
And Table \ref{tab:dataset_info} provides a summary of the basic information for all datasets we analyze.

\begin{table*}[h!]
	\centering
	\caption{Benchmark dataset links.}
	\label{sup:tab:dataset_urls}
	\resizebox{0.75\textwidth}{!}{
		\begin{tabular}{l | l}
			\toprule
			Dataset Name &          URL \\
			\midrule
			\texttt{adult} &         \url{https://www.kaggle.com/lodetomasi1995/income-classification} \\
			\texttt{arcene} &        \url{https://www.openml.org/search?type=data&sort=runs&id=1458} \\
			\texttt{arrhythmia} &    \url{https://archive.ics.uci.edu/dataset/5/arrhythmia} \\
			\texttt{bank}  &         \url{https://archive.ics.uci.edu/ml/datasets/bank+marketing} \\
			\texttt{blastChar} &     \url{https://www.kaggle.com/blastchar/telco-customer-churn} \\
			\texttt{churn} &         \url{https://www.kaggle.com/shrutimechlearn/churn-modelling} \\
			\texttt{htru2} &         \url{https://archive.ics.uci.edu/ml/datasets/HTRU2} \\
			\texttt{qsar\_bio} &     \url{https://archive.ics.uci.edu/ml/datasets/QSAR+biodegradation} \\
			\texttt{shoppers} &      \url{https://archive.ics.uci.edu/ml/datasets/Online+Shoppers+Purchasing+Intention+Dataset} \\
			\texttt{spambase} &      \url{https://archive.ics.uci.edu/ml/datasets/Spambase} \\    
			\midrule
	\end{tabular}}
\end{table*}

\begin{table*}[h!]
	\caption{Description of benchmark datasets.  
    All datasets used in our experiments are binary classification tasks.
    ``Positive Class ratio (\%)" indicates the proportion of samples labeled as belonging to the positive class.}
	\label{tab:dataset_info}
	\centering
	\scalebox{0.9}{
		\begin{tabular}{l | ccccc}
			\toprule
			\multicolumn{1}{c|}{\textbf{Dataset}} & \textbf{Dataset size} & \textbf{\# Features} & \textbf{\# Categorical} & \textbf{\# Continuous} & \textbf{Positive class ratio (\%)} \\ 
			\midrule
			\texttt{adult}      & 32561 & 14  & 8  & 6   & 24.08 \\
			\texttt{arcene}     & 200   & 783 & 0  & 783 & 44.00    \\
			\texttt{arrhythmia} & 452   & 226 & 0  & 226 & 14.60  \\
			\texttt{bank}       & 45211 & 16  & 9  & 7   & 11.70  \\
			\texttt{blastChar}  & 7043  & 20  & 17 & 3   & 26.54 \\
			\texttt{churn}      & 10000 & 11  & 3  & 8   & 20.37 \\
			\texttt{htru2}      & 17898 & 8   & 0  & 8   & 9.16  \\
			\texttt{qsar\_bio}  & 1055  & 41  & 0  & 41  & 33.74 \\
			\texttt{shoppers}   & 12330 & 17  & 2  & 15  & 15.47 \\
			\texttt{spambase}   & 4601  & 57  & 0  & 57  & 39.40  \\
			\midrule
	\end{tabular}}
\end{table*}

\subsection{B-\MakeUppercase{\romannumeral 2}. Missingness Generation Mechanism}
\label{sec:appB2_miss}

Here, we describe the missingness generation process for the three types of missingness mechanisms: MCAR, MAR, and MNAR. 
We follow a similar approach to that of \citet{rockenschaub2024robust} to impose missingness.
Prior to imposing missingness, we convert all categorical variables into continuous ones by mapping each category to a pre-specified integer.
For example, a categorical variable with four distinct values is encoded as integers ranging from 0 to 3.
After this transformation, we apply standardization so that each variable has zero mean and unit variance.

Let $x_{ij}$ denote the real value and $m_{ij}$ the corresponding masking indicator for the $i$-th sample and $j$-th variable.
Note that $m_{ij} = 1$ indicates that the value $x_{ij}$ is missing.

\paragraph{MCAR} 
For the $j$-th variable and $i$-th sample, we draw a random number $u_{ij}$ from the uniform distribution $U(0,1)$ and determine the masking indicator $m_{ij}$ as follows:
\begin{align*}
m_{ij} = \mathbb{I}(u_{ij}<\alpha).
\end{align*}

\paragraph{MAR}
First, we randomly select a subset $\mathcal{J}$ consisting of 30\% of the variable indices, and apply the MCAR process to each variable $x_j$ with $j \in \mathcal{J}$, using a missingness probability of $\alpha$. 
For the $i$-th sample, let $\boldsymbol{x}_{i\mathcal{J}}^0$ denote the zero-imputed vector of $x_{i\mathcal{J}}$.
We also let $\boldsymbol{x}_{i\mathcal{J}^c}$ denote the complementary vector, corresponding to the features not included in $\mathcal{J}$.

For a given variable $x_{j}$ for $j\in \mathcal{J}^c$, we draw a $|\mathcal{J}|$-dimensional random vector $\boldsymbol{\gamma}_{j}$ from $\mathcal{N}(\boldsymbol{0}_{|\mathcal{J}|}, \boldsymbol{I}_{|\mathcal{J}|})$. 
Then, for each $i$-th sample, we compute:
\begin{align*}
\alpha_{ij} = \delta_j \cdot \sigma(\boldsymbol{\gamma}_{j}^T \boldsymbol{x}_{i\mathcal{J}}^0),
\end{align*}
where $\sigma(\cdot)$ is the sigmoid function and $\delta_j$ is a rescaling factor chosen such that the mean of $\alpha_{ij}$ across all $i \in [n]$ equals the target missingness probability $\alpha$. 
Then, with a random number $u_{ij}\sim U(0,1)$, the masking indicator $m_{ij}$ is calculated as:
\begin{align*}
m_{ij} = \mathbb{I}(u_{ij}<\alpha_{ij}).    
\end{align*}

\paragraph{MNAR}
Similar to the MAR scenario, we begin by randomly selecting 30\% of the variable indices to form a subset $\mathcal{J}$, and apply the MCAR mechanism to each variable $x_j$ for $j \in \mathcal{J}$ with a missingness probability of $\alpha$.
For each $i$-th sample, let $\boldsymbol{x}_i^0$ denote the zero-imputed version of $\boldsymbol{x}_i$ obtained after applying the MCAR process described above.

For a given variable $x_{j}$ for $j\in \mathcal{J}^c$, we draw a $p$-dimensional random vector $\boldsymbol{\gamma}_{j}$ from $\mathcal{N}(\boldsymbol{0}_p, \boldsymbol{I}_p)$. 
Then, for each $i$-th sample, we calculate:
\begin{align*}
\alpha_{ij} = \delta_j \cdot \sigma(\boldsymbol{\gamma}_{j}^T \boldsymbol{x}_i^0),
\end{align*}
where $\delta_j$ is a normalization constant chosen so that the average of $\alpha_{ij}$ over all $i \in [n]$ matches the target missingness probability $\alpha$.
Next, given a random draw $u_{ij} \sim U(0,1)$, the masking indicator $m_{ij}$ is computed as:
\begin{align*}
m_{ij} = \mathbb{I}(u_{ij}<\alpha_{ij}).    
\end{align*}

\subsection{B-\MakeUppercase{\romannumeral 3}. Architecture Description}
\label{sec:appB3}
Regarding the embedding mechanism, we employ a single-hidden-layer MLP with 100 hidden units for each continuous variable.
For categorical variables, each distinct value is treated as a token and mapped to a learnable embedding vector.
Additionally, for each variable, we introduce a separate learnable embedding vector to represent missing values.
Notably, a distinct embedding is assigned for the missing case of each variable, allowing the model to capture variable-specific missingness.

We employed a Transformer-based architecture to model the tabular data.
By default, the Transformer encoder depth was set to 6, the number of attention heads to 8, and the embedding dimension to 32. 
For high-dimensional datasets with more than 100 input features, the embedding dimension was reduced to 4 and the batch size was  restricted to 64. 
Furthermore, for \texttt{Arrhythmia} dataset, the encoder depth was reduced to 1 to lower the risk of overfitting, and for \texttt{Arcene} dataset, the depth and number of heads were respectively adjusted to 4 and 1 in order to tailor the model capacity to the limited sample size.

\subsection{B-\MakeUppercase{\romannumeral 4}. Detailed Results for Input-Missingness Scenarios}
\label{sec:appB4}
Tables \ref{sup:tab:adult_auc}-\ref{sup:tab:spambase_std} present the detailed results of average AUC scores and their standard deviations, computed over three independent runs of each method across all benchmark datasets under the input-missingness scenarios.

\begin{sidewaystable}[htbp]
\renewcommand\thetable{B.3}
\centering
\caption{Comparison of averaged test AUC scores (\%) on \texttt{adult} for various $(\alpha^{\text{tr}},\alpha^{\text{ts}})$ combinations.}
\label{sup:tab:adult_auc}
\fontsize{6pt}{7pt}\selectfont
\resizebox{0.99\textwidth}{!}{
\begin{tabular}{l|c|cccccccc|cccccccc|cccccccc}
\hline
\textbf{Method} & \textbf{No Missing} & \multicolumn{8}{c|}{\textbf{MAR}} & \multicolumn{8}{c|}{\textbf{MNAR}} & \multicolumn{8}{c}{\textbf{MNAR}} \\ \hline
$\alpha^{\text{tr}}$ & 0 & \multicolumn{2}{c|}{0} & \multicolumn{3}{c|}{0.15} & \multicolumn{3}{c|}{0.3} & \multicolumn{2}{c|}{0} & \multicolumn{3}{c|}{0.15} & \multicolumn{3}{c|}{0.3} & \multicolumn{2}{c|}{0} & \multicolumn{3}{c|}{0.15} & \multicolumn{3}{c}{0.3} \\ \hline
$\alpha^{\text{ts}}$ & 0 & 0.15 & \multicolumn{1}{c|}{0.3} & 0 & 0.15 & \multicolumn{1}{c|}{0.3} & 0 & 0.15 & 0.3 & 0.15 & \multicolumn{1}{c|}{0.3} & 0 & 0.15 & \multicolumn{1}{c|}{0.3} & 0 & 0.15 & 0.3 & 0.15 & \multicolumn{1}{c|}{0.3} & 0 & 0.15 & \multicolumn{1}{c|}{0.3} & 0 & 0.15 & 0.3 \\ \hline
LR-0 & 60.90 & 58.90 & \multicolumn{1}{c|}{56.73} & 60.53 & 58.65 & \multicolumn{1}{c|}{56.57} & 60.11 & 58.36 & 56.39 & 58.65 & \multicolumn{1}{c|}{56.52} & 60.05 & 58.07 & \multicolumn{1}{c|}{56.16} & 59.71 & 57.80 & 55.99 & 58.79 & \multicolumn{1}{c|}{56.66} & 60.33 & 58.42 & \multicolumn{1}{c|}{56.39} & 59.93 & 58.14 & 56.18 \\
NN-0 & 51.19 & 50.97 & \multicolumn{1}{c|}{51.40} & 53.00 & 59.74 & \multicolumn{1}{c|}{63.66} & 50.29 & 57.33 & 62.07 & 51.05 & \multicolumn{1}{c|}{52.18} & 49.14 & 55.16 & \multicolumn{1}{c|}{59.18} & 48.49 & 55.07 & 60.28 & 51.08 & \multicolumn{1}{c|}{51.89} & 49.27 & 57.63 & \multicolumn{1}{c|}{62.21} & 48.26 & 55.51 & 60.78 \\
RF & 90.84 & 88.70 & \multicolumn{1}{c|}{86.44} & 90.15 & 88.85 & \multicolumn{1}{c|}{87.16} & 89.71 & 88.74 & 87.10 & 88.70 & \multicolumn{1}{c|}{86.11} & 90.21 & 88.88 & \multicolumn{1}{c|}{86.86} & 89.95 & 88.87 & 87.25 & 88.87 & \multicolumn{1}{c|}{85.99} & 90.29 & 89.23 & \multicolumn{1}{c|}{86.97} & 90.13 & 89.10 & 87.33 \\
\hline
\end{tabular}}

\vspace{0.3cm}

\renewcommand\thetable{B.4}
\centering
\caption{Standard deviation comparison of test AUC scores (\%) on \texttt{adult} for various $(\alpha^{\text{tr}}, \alpha^{\text{ts}})$ combinations.}
\label{sup:tab:adult_std}
\fontsize{6pt}{7pt}\selectfont
\resizebox{0.99\textwidth}{!}{
\begin{tabular}{l|c|cccccccc|cccccccc|cccccccc}
\hline
\textbf{Method} & \textbf{No Missing} & \multicolumn{8}{c|}{\textbf{MAR}} & \multicolumn{8}{c|}{\textbf{MNAR}} & \multicolumn{8}{c}{\textbf{MCAR}} \\ \hline
$\alpha^{\text{tr}}$ & 0 & \multicolumn{2}{c|}{0} & \multicolumn{3}{c|}{0.15} & \multicolumn{3}{c|}{0.3} & \multicolumn{2}{c|}{0} & \multicolumn{3}{c|}{0.15} & \multicolumn{3}{c|}{0.3} & \multicolumn{2}{c|}{0} & \multicolumn{3}{c|}{0.15} & \multicolumn{3}{c}{0.3} \\ \hline
$\alpha^{\text{ts}}$ & 0 & 0.15 & \multicolumn{1}{c|}{0.3} & 0 & 0.15 & \multicolumn{1}{c|}{0.3} & 0 & 0.15 & 0.3 & 0.15 & \multicolumn{1}{c|}{0.3} & 0 & 0.15 & \multicolumn{1}{c|}{0.3} & 0 & 0.15 & 0.3 & 0.15 & \multicolumn{1}{c|}{0.3} & 0 & 0.15 & \multicolumn{1}{c|}{0.3} & 0 & 0.15 & 0.3 \\ \hline
LR-0 & 0.91 & 1.11 & \multicolumn{1}{c|}{1.27} & 1.09 & 1.15 & \multicolumn{1}{c|}{1.33} & 0.82 & 1.04 & 1.22 & 0.39 & \multicolumn{1}{c|}{0.25} & 0.49 & 0.21 & \multicolumn{1}{c|}{0.26} & 0.85 & 0.50 & 0.37 & 0.33 & \multicolumn{1}{c|}{0.40} & 0.62 & 0.27 & \multicolumn{1}{c|}{0.32} & 1.03 & 0.20 & 0.24 \\
NN-0 & 2.35 & 0.68 & \multicolumn{1}{c|}{1.00} & 3.90 & 2.41 & \multicolumn{1}{c|}{1.74} & 0.58 & 1.37 & 2.15 & 0.56 & \multicolumn{1}{c|}{1.19} & 1.02 & 1.28 & \multicolumn{1}{c|}{2.39} & 1.39 & 2.85 & 4.08 & 1.10 & \multicolumn{1}{c|}{1.15} & 0.72 & 1.21 & \multicolumn{1}{c|}{1.13} & 2.28 & 2.12 & 2.90 \\
RF & 0.26 & 0.61 & \multicolumn{1}{c|}{0.54} & 0.62 & 0.49 & \multicolumn{1}{c|}{0.65} & 0.61 & 0.69 & 0.66 & 0.27 & \multicolumn{1}{c|}{0.49} & 0.43 & 0.27 & \multicolumn{1}{c|}{0.53} & 0.52 & 0.43 & 0.42 & 0.46 & \multicolumn{1}{c|}{0.70} & 0.42 & 0.42 & \multicolumn{1}{c|}{0.64} & 0.52 & 0.55 & 0.48 \\
\hline
\end{tabular}}

\vspace{0.3cm}

\renewcommand\thetable{B.5}
\centering
\caption{Comparison of averaged test AUC scores (\%) on \texttt{arcene} for various $(\alpha^{\text{tr}},\alpha^{\text{ts}})$ combinations.}
\label{sup:tab:arcene_auc}
\fontsize{6pt}{7pt}\selectfont
\resizebox{0.99\textwidth}{!}{
\begin{tabular}{l|c|cccccccc|cccccccc|cccccccc}
\hline
\textbf{Method} & \textbf{No Missing} & \multicolumn{8}{c|}{\textbf{MAR}} & \multicolumn{8}{c|}{\textbf{MNAR}} & \multicolumn{8}{c}{\textbf{MCAR}} \\ \hline
$\alpha^{\text{tr}}$ & 0 & \multicolumn{2}{c|}{0} & \multicolumn{3}{c|}{0.15} & \multicolumn{3}{c|}{0.3} & \multicolumn{2}{c|}{0} & \multicolumn{3}{c|}{0.15} & \multicolumn{3}{c|}{0.3} & \multicolumn{2}{c|}{0} & \multicolumn{3}{c|}{0.15} & \multicolumn{3}{c}{0.3} \\ \hline
$\alpha^{\text{ts}}$ & 0 & 0.15 & \multicolumn{1}{c|}{0.3} & 0 & 0.15 & \multicolumn{1}{c|}{0.3} & 0 & 0.15 & 0.3 & 0.15 & \multicolumn{1}{c|}{0.3} & 0 & 0.15 & \multicolumn{1}{c|}{0.3} & 0 & 0.15 & 0.3 & 0.15 & \multicolumn{1}{c|}{0.3} & 0 & 0.15 & \multicolumn{1}{c|}{0.3} & 0 & 0.15 & 0.3 \\ \hline
LR-0 & 63.89 & 64.04 & \multicolumn{1}{c|}{61.57} & 58.80 & 58.10 & \multicolumn{1}{c|}{55.86} & 59.80 & 59.26 & 56.87 & 64.51 & \multicolumn{1}{c|}{64.81} & 59.80 & 58.56 & \multicolumn{1}{c|}{60.80} & 60.19 & 57.25 & 59.10 & 66.20 & \multicolumn{1}{c|}{66.28} & 58.80 & 59.57 & \multicolumn{1}{c|}{59.72} & 59.18 & 56.48 & 58.87 \\
NN-0 & 49.31 & 61.81 & \multicolumn{1}{c|}{53.97} & 76.08 & 75.62 & \multicolumn{1}{c|}{73.42} & 78.09 & 77.70 & 76.77 & 60.34 & \multicolumn{1}{c|}{57.10} & 76.31 & 69.17 & \multicolumn{1}{c|}{67.67} & 74.54 & 69.44 & 66.47 & 63.31 & \multicolumn{1}{c|}{59.88} & 74.38 & 71.14 & \multicolumn{1}{c|}{66.40} & 79.51 & 76.70 & 76.20 \\
RF & 50.66 & 61.23 & \multicolumn{1}{c|}{63.54} & 62.62 & 66.94 & \multicolumn{1}{c|}{61.42} & 58.18 & 61.19 & 56.33 & 66.55 & \multicolumn{1}{c|}{57.95} & 64.00 & 66.36 & \multicolumn{1}{c|}{64.51} & 63.73 & 59.53 & 60.73 & 62.89 & \multicolumn{1}{c|}{64.16} & 60.49 & 58.99 & \multicolumn{1}{c|}{57.68} & 59.49 & 62.31 & 56.94 \\
\hline
\end{tabular}
}

\vspace{0.3cm}

\renewcommand\thetable{B.6}
\centering
\caption{Standard deviation comparison of test AUC scores (\%) on \texttt{arcene} for various $(\alpha^{\text{tr}}, \alpha^{\text{ts}})$ combinations.}
\label{sup:tab:arcene_std}
\fontsize{6pt}{7pt}\selectfont
\resizebox{0.99\textwidth}{!}{
\begin{tabular}{l|c|cccccccc|cccccccc|cccccccc}
\hline
\textbf{Method} & \textbf{No Missing} & \multicolumn{8}{c|}{\textbf{MAR}} & \multicolumn{8}{c|}{\textbf{MNAR}} & \multicolumn{8}{c}{\textbf{MCAR}} \\ \hline
$\alpha^{\text{tr}}$ & 0 & \multicolumn{2}{c|}{0} & \multicolumn{3}{c|}{0.15} & \multicolumn{3}{c|}{0.3} & \multicolumn{2}{c|}{0} & \multicolumn{3}{c|}{0.15} & \multicolumn{3}{c|}{0.3} & \multicolumn{2}{c|}{0} & \multicolumn{3}{c|}{0.15} & \multicolumn{3}{c}{0.3} \\ \hline
$\alpha^{\text{ts}}$ & 0 & 0.15 & \multicolumn{1}{c|}{0.3} & 0 & 0.15 & \multicolumn{1}{c|}{0.3} & 0 & 0.15 & 0.3 & 0.15 & \multicolumn{1}{c|}{0.3} & 0 & 0.15 & \multicolumn{1}{c|}{0.3} & 0 & 0.15 & 0.3 & 0.15 & \multicolumn{1}{c|}{0.3} & 0 & 0.15 & \multicolumn{1}{c|}{0.3} & 0 & 0.15 & 0.3 \\ \hline
LR-0 & 14.21 & 6.47 & \multicolumn{1}{c|}{5.03} & 6.93 & 7.04 & \multicolumn{1}{c|}{7.69} & 5.80 & 3.30 & 4.85 & 4.68 & \multicolumn{1}{c|}{3.21} & 7.09 & 4.78 & \multicolumn{1}{c|}{3.60} & 6.47 & 9.57 & 7.89 & 1.82 & \multicolumn{1}{c|}{7.49} & 3.45 & 7.08 & \multicolumn{1}{c|}{10.69} & 7.55 & 8.78 & 11.23 \\
NN-0 & 1.21 & 12.22 & \multicolumn{1}{c|}{13.74} & 17.80 & 17.12 & \multicolumn{1}{c|}{17.74} & 18.61 & 16.89 & 17.11 & 12.80 & \multicolumn{1}{c|}{10.28} & 17.91 & 16.74 & \multicolumn{1}{c|}{15.59} & 18.99 & 18.63 & 16.72 & 10.93 & \multicolumn{1}{c|}{12.25} & 17.19 & 16.21 & \multicolumn{1}{c|}{15.82} & 18.45 & 15.86 & 17.03 \\
RF & 17.68 & 15.58 & \multicolumn{1}{c|}{14.56} & 17.56 & 18.61 & \multicolumn{1}{c|}{17.61} & 16.44 & 17.07 & 14.26 & 18.01 & \multicolumn{1}{c|}{17.72} & 15.18 & 13.41 & \multicolumn{1}{c|}{14.14} & 13.72 & 14.24 & 14.69 & 14.95 & \multicolumn{1}{c|}{16.96} & 14.94 & 12.64 & \multicolumn{1}{c|}{12.51} & 10.87 & 12.71 & 14.15 \\
\hline
\end{tabular}
}
\end{sidewaystable}

\begin{sidewaystable}[htbp]
\renewcommand\thetable{B.7}
\centering
\caption{Comparison of averaged test AUC scores (\%) on \texttt{arrhythmia} for various $(\alpha^{\text{tr}},\alpha^{\text{ts}})$ combinations.}
\label{sup:tab:arrhythmia_auc}
\fontsize{6pt}{7pt}\selectfont
\resizebox{0.99\textwidth}{!}{
\begin{tabular}{l|c|cccccccc|cccccccc|cccccccc}
\hline
\textbf{Method} & \textbf{No Missing} & \multicolumn{8}{c|}{\textbf{MAR}} & \multicolumn{8}{c|}{\textbf{MNAR}} & \multicolumn{8}{c}{\textbf{MCAR}} \\ \hline
$\alpha^{\text{tr}}$ & 0 & \multicolumn{2}{c|}{0} & \multicolumn{3}{c|}{0.15} & \multicolumn{3}{c|}{0.3} & \multicolumn{2}{c|}{0} & \multicolumn{3}{c|}{0.15} & \multicolumn{3}{c|}{0.3} & \multicolumn{2}{c|}{0} & \multicolumn{3}{c|}{0.15} & \multicolumn{3}{c}{0.3} \\ \hline
$\alpha^{\text{ts}}$ & 0 & 0.15 & \multicolumn{1}{c|}{0.3} & 0 & 0.15 & \multicolumn{1}{c|}{0.3} & 0 & 0.15 & 0.3 & 0.15 & \multicolumn{1}{c|}{0.3} & 0 & 0.15 & \multicolumn{1}{c|}{0.3} & 0 & 0.15 & 0.3 & 0.15 & \multicolumn{1}{c|}{0.3} & 0 & 0.15 & \multicolumn{1}{c|}{0.3} & 0 & 0.15 & 0.3 \\ \hline
LR-0 & 64.50 & 64.87 & \multicolumn{1}{c|}{68.07} & 69.73 & 72.52 & \multicolumn{1}{c|}{69.99} & 71.98 & 71.20 & 66.64 & 64.35 & \multicolumn{1}{c|}{64.22} & 68.52 & 67.00 & \multicolumn{1}{c|}{70.30} & 69.70 & 70.07 & 69.79 & 67.48 & \multicolumn{1}{c|}{66.27} & 66.30 & 65.77 & \multicolumn{1}{c|}{61.69} & 63.38 & 64.64 & 63.97 \\
NN-0 & 76.99 & 75.67 & \multicolumn{1}{c|}{75.06} & 75.16 & 71.91 & \multicolumn{1}{c|}{73.64} & 74.77 & 70.38 & 73.68 & 72.69 & \multicolumn{1}{c|}{69.23} & 80.98 & 81.28 & \multicolumn{1}{c|}{77.22} & 78.26 & 77.67 & 78.28 & 76.98 & \multicolumn{1}{c|}{67.78} & 80.24 & 73.77 & \multicolumn{1}{c|}{72.42} & 76.13 & 69.90 & 67.12 \\
RF & 73.14 & 80.03 & \multicolumn{1}{c|}{78.76} & 81.72 & 80.73 & \multicolumn{1}{c|}{81.46} & 79.94 & 81.93 & 68.66 & 81.21 & \multicolumn{1}{c|}{77.27} & 82.45 & 82.48 & \multicolumn{1}{c|}{83.80} & 74.85 & 76.37 & 74.22 & 80.39 & \multicolumn{1}{c|}{80.65} & 79.93 & 81.80 & \multicolumn{1}{c|}{80.04} & 80.45 & 81.72 & 75.47 \\
\hline
\end{tabular}
}

\vspace{0.3cm}

\renewcommand\thetable{B.8}
\centering
\caption{Standard deviation comparison of test AUC scores (\%) on \texttt{arrhythmia} for various $(\alpha^{\text{tr}}, \alpha^{\text{ts}})$ combinations.}
\label{sup:tab:arrhythmia_std}
\fontsize{6pt}{7pt}\selectfont
\resizebox{0.99\textwidth}{!}{
\begin{tabular}{l|c|cccccccc|cccccccc|cccccccc}
\hline
\textbf{Method} & \textbf{No Missing} & \multicolumn{8}{c|}{\textbf{MAR}} & \multicolumn{8}{c|}{\textbf{MNAR}} & \multicolumn{8}{c}{\textbf{MCAR}} \\ \hline
$\alpha^{\text{tr}}$ & 0 & \multicolumn{2}{c|}{0} & \multicolumn{3}{c|}{0.15} & \multicolumn{3}{c|}{0.3} & \multicolumn{2}{c|}{0} & \multicolumn{3}{c|}{0.15} & \multicolumn{3}{c|}{0.3} & \multicolumn{2}{c|}{0} & \multicolumn{3}{c|}{0.15} & \multicolumn{3}{c}{0.3} \\ \hline
$\alpha^{\text{ts}}$ & 0 & 0.15 & \multicolumn{1}{c|}{0.3} & 0 & 0.15 & \multicolumn{1}{c|}{0.3} & 0 & 0.15 & 0.3 & 0.15 & \multicolumn{1}{c|}{0.3} & 0 & 0.15 & \multicolumn{1}{c|}{0.3} & 0 & 0.15 & 0.3 & 0.15 & \multicolumn{1}{c|}{0.3} & 0 & 0.15 & \multicolumn{1}{c|}{0.3} & 0 & 0.15 & 0.3 \\ \hline
LR-0 & 10.66 & 7.30 & \multicolumn{1}{c|}{4.03} & 11.34 & 13.38 & \multicolumn{1}{c|}{8.21} & 9.48 & 7.15 & 6.47 & 9.76 & \multicolumn{1}{c|}{6.62} & 11.91 & 11.51 & \multicolumn{1}{c|}{8.62} & 9.30 & 8.21 & 10.30 & 8.92 & \multicolumn{1}{c|}{8.80} & 9.34 & 3.39 & \multicolumn{1}{c|}{1.52} & 0.79 & 5.31 & 4.52 \\
NN-0 & 8.93 & 7.76 & \multicolumn{1}{c|}{4.84} & 8.30 & 6.65 & \multicolumn{1}{c|}{5.42} & 6.80 & 4.59 & 4.90 & 10.71 & \multicolumn{1}{c|}{10.53} & 5.49 & 3.93 & \multicolumn{1}{c|}{7.14} & 6.48 & 4.16 & 1.69 & 9.73 & \multicolumn{1}{c|}{11.34} & 6.61 & 6.48 & \multicolumn{1}{c|}{9.35} & 10.70 & 7.40 & 7.73 \\
RF & 10.53 & 12.58 & \multicolumn{1}{c|}{9.54} & 10.64 & 10.07 & \multicolumn{1}{c|}{10.53} & 11.07 & 10.27 & 8.80 & 11.06 & \multicolumn{1}{c|}{9.05} & 7.80 & 6.06 & \multicolumn{1}{c|}{5.14} & 11.50 & 10.22 & 10.19 & 11.92 & \multicolumn{1}{c|}{10.14} & 11.33 & 8.26 & \multicolumn{1}{c|}{9.32} & 9.51 & 8.52 & 5.47 \\
\hline
\end{tabular}
}

\vspace{0.3cm}

\renewcommand\thetable{B.9}
\centering
\caption{Comparison of averaged test AUC scores (\%) on \texttt{bank} for various $(\alpha^{\text{tr}},\alpha^{\text{ts}})$ combinations.}
\label{sup:tab:bank_auc}
\fontsize{6pt}{7pt}\selectfont
\resizebox{0.99\textwidth}{!}{
\begin{tabular}{l|c|cccccccc|cccccccc|cccccccc}
\hline
\textbf{Method} & \textbf{No Missing} & \multicolumn{8}{c|}{\textbf{MAR}} & \multicolumn{8}{c|}{\textbf{MNAR}} & \multicolumn{8}{c}{\textbf{MCAR}} \\ \hline
$\alpha^{\text{tr}}$ & 0 & \multicolumn{2}{c|}{0} & \multicolumn{3}{c|}{0.15} & \multicolumn{3}{c|}{0.3} & \multicolumn{2}{c|}{0} & \multicolumn{3}{c|}{0.15} & \multicolumn{3}{c|}{0.3} & \multicolumn{2}{c|}{0} & \multicolumn{3}{c|}{0.15} & \multicolumn{3}{c}{0.3} \\ \hline
$\alpha^{\text{ts}}$ & 0 & 0.15 & \multicolumn{1}{c|}{0.3} & 0 & 0.15 & \multicolumn{1}{c|}{0.3} & 0 & 0.15 & 0.3 & 0.15 & \multicolumn{1}{c|}{0.3} & 0 & 0.15 & \multicolumn{1}{c|}{0.3} & 0 & 0.15 & 0.3 & 0.15 & \multicolumn{1}{c|}{0.3} & 0 & 0.15 & \multicolumn{1}{c|}{0.3} & 0 & 0.15 & 0.3 \\ \hline
LR-0 & 69.12 & 63.58 & \multicolumn{1}{c|}{59.21} & 62.74 & 59.27 & \multicolumn{1}{c|}{56.15} & 54.33 & 53.89 & 52.23 & 63.06 & \multicolumn{1}{c|}{58.91} & 59.24 & 56.63 & \multicolumn{1}{c|}{54.44} & 50.18 & 50.67 & 50.27 & 62.92 & \multicolumn{1}{c|}{59.19} & 59.13 & 56.14 & \multicolumn{1}{c|}{54.51} & 48.06 & 48.90 & 49.48 \\
NN-0 & 83.20 & 75.05 & \multicolumn{1}{c|}{68.18} & 82.55 & 77.02 & \multicolumn{1}{c|}{71.19} & 82.81 & 79.33 & 75.02 & 74.48 & \multicolumn{1}{c|}{67.29} & 82.47 & 76.54 & \multicolumn{1}{c|}{70.63} & 83.60 & 79.86 & 75.37 & 74.79 & \multicolumn{1}{c|}{67.73} & 82.52 & 76.81 & \multicolumn{1}{c|}{71.19} & 82.85 & 79.07 & 74.96 \\
RF & 90.45 & 86.04 & \multicolumn{1}{c|}{81.11} & 89.66 & 87.14 & \multicolumn{1}{c|}{83.98} & 89.54 & 86.93 & 83.94 & 85.67 & \multicolumn{1}{c|}{81.13} & 89.80 & 87.02 & \multicolumn{1}{c|}{83.84} & 89.34 & 86.66 & 83.92 & 85.92 & \multicolumn{1}{c|}{81.00} & 89.76 & 87.07 & \multicolumn{1}{c|}{84.03} & 89.06 & 86.81 & 84.05 \\
\hline
\end{tabular}
}

\vspace{0.3cm}

\renewcommand\thetable{B.10}
\centering
\caption{Standard deviation comparison of test AUC scores (\%) on \texttt{bank} for various $(\alpha^{\text{tr}}, \alpha^{\text{ts}})$ combinations.}
\label{sup:tab:bank_std}

\fontsize{6pt}{7pt}\selectfont
\resizebox{0.99\textwidth}{!}{
\begin{tabular}{l|c|cccccccc|cccccccc|cccccccc}
\hline
\textbf{Method} & \textbf{No Missing} & \multicolumn{8}{c|}{\textbf{MAR}} & \multicolumn{8}{c|}{\textbf{MNAR}} & \multicolumn{8}{c}{\textbf{MCAR}} \\ \hline
$\alpha^{\text{tr}}$ & 0 & \multicolumn{2}{c|}{0} & \multicolumn{3}{c|}{0.15} & \multicolumn{3}{c|}{0.3} & \multicolumn{2}{c|}{0} & \multicolumn{3}{c|}{0.15} & \multicolumn{3}{c|}{0.3} & \multicolumn{2}{c|}{0} & \multicolumn{3}{c|}{0.15} & \multicolumn{3}{c}{0.3} \\ \hline
$\alpha^{\text{ts}}$ & 0 & 0.15 & \multicolumn{1}{c|}{0.3} & 0 & 0.15 & \multicolumn{1}{c|}{0.3} & 0 & 0.15 & 0.3 & 0.15 & \multicolumn{1}{c|}{0.3} & 0 & 0.15 & \multicolumn{1}{c|}{0.3} & 0 & 0.15 & 0.3 & 0.15 & \multicolumn{1}{c|}{0.3} & 0 & 0.15 & \multicolumn{1}{c|}{0.3} & 0 & 0.15 & 0.3 \\ \hline
LR-0 & 1.01 & 0.84 & \multicolumn{1}{c|}{0.33} & 2.92 & 2.05 & \multicolumn{1}{c|}{0.99} & 5.30 & 3.23 & 2.00 & 0.85 & \multicolumn{1}{c|}{0.90} & 2.25 & 0.63 & \multicolumn{1}{c|}{0.20} & 2.23 & 1.01 & 1.01 & 0.83 & \multicolumn{1}{c|}{1.21} & 1.81 & 1.53 & \multicolumn{1}{c|}{1.51} & 4.12 & 2.99 & 2.51 \\
NN-0 & 0.56 & 0.23 & \multicolumn{1}{c|}{0.46} & 0.60 & 0.54 & \multicolumn{1}{c|}{1.09} & 1.20 & 1.38 & 2.25 & 0.42 & \multicolumn{1}{c|}{0.60} & 1.13 & 1.18 & \multicolumn{1}{c|}{1.46} & 2.04 & 2.08 & 2.70 & 0.33 & \multicolumn{1}{c|}{0.18} & 1.05 & 0.79 & \multicolumn{1}{c|}{1.01} & 2.17 & 2.02 & 1.99 \\
RF & 0.13 & 0.88 & \multicolumn{1}{c|}{1.98} & 0.18 & 0.21 & \multicolumn{1}{c|}{0.67} & 0.44 & 0.37 & 0.95 & 0.82 & \multicolumn{1}{c|}{0.49} & 0.16 & 0.21 & \multicolumn{1}{c|}{0.63} & 0.20 & 0.39 & 0.56 & 0.81 & \multicolumn{1}{c|}{0.92} & 0.27 & 0.18 & \multicolumn{1}{c|}{0.38} & 0.38 & 0.39 & 0.54 \\
\hline
\end{tabular}
}
\end{sidewaystable}

\begin{sidewaystable}[htbp]
\renewcommand\thetable{B.11}
\centering
\caption{Comparison of averaged test AUC scores (\%) on \texttt{blastchar} for various $(\alpha^{\text{tr}},\alpha^{\text{ts}})$ combinations.}
\label{sup:tab:blastchar_auc}
\fontsize{6pt}{7pt}\selectfont
\resizebox{0.99\textwidth}{!}{
\begin{tabular}{l|c|cccccccc|cccccccc|cccccccc}
\hline
\textbf{Method} & \textbf{No Missing} & \multicolumn{8}{c|}{\textbf{MAR}} & \multicolumn{8}{c|}{\textbf{MNAR}} & \multicolumn{8}{c}{\textbf{MCAR}} \\ \hline
$\alpha^{\text{tr}}$ & 0 & \multicolumn{2}{c|}{0} & \multicolumn{3}{c|}{0.15} & \multicolumn{3}{c|}{0.3} & \multicolumn{2}{c|}{0} & \multicolumn{3}{c|}{0.15} & \multicolumn{3}{c|}{0.3} & \multicolumn{2}{c|}{0} & \multicolumn{3}{c|}{0.15} & \multicolumn{3}{c}{0.3} \\ \hline
$\alpha^{\text{ts}}$ & 0 & 0.15 & \multicolumn{1}{c|}{0.3} & 0 & 0.15 & \multicolumn{1}{c|}{0.3} & 0 & 0.15 & 0.3 & 0.15 & \multicolumn{1}{c|}{0.3} & 0 & 0.15 & \multicolumn{1}{c|}{0.3} & 0 & 0.15 & 0.3 & 0.15 & \multicolumn{1}{c|}{0.3} & 0 & 0.15 & \multicolumn{1}{c|}{0.3} & 0 & 0.15 & 0.3 \\ \hline
LR-0 & 79.52 & 71.03 & \multicolumn{1}{c|}{63.70} & 75.60 & 72.16 & \multicolumn{1}{c|}{69.00} & 72.90 & 70.76 & 68.30 & 71.82 & \multicolumn{1}{c|}{65.90} & 75.66 & 72.80 & \multicolumn{1}{c|}{69.31} & 73.70 & 72.01 & 69.03 & 71.27 & \multicolumn{1}{c|}{64.14} & 74.59 & 71.55 & \multicolumn{1}{c|}{67.63} & 73.01 & 70.77 & 67.74 \\
NN-0 & 80.74 & 71.53 & \multicolumn{1}{c|}{65.62} & 82.84 & 80.98 & \multicolumn{1}{c|}{78.51} & 82.35 & 80.71 & 78.99 & 70.54 & \multicolumn{1}{c|}{64.23} & 83.34 & 81.30 & \multicolumn{1}{c|}{78.41} & 78.20 & 77.73 & 76.61 & 71.03 & \multicolumn{1}{c|}{65.42} & 83.24 & 81.32 & \multicolumn{1}{c|}{78.11} & 81.10 & 79.80 & 77.60 \\
RF & 83.68 & 82.60 & \multicolumn{1}{c|}{81.38} & 83.45 & 82.52 & \multicolumn{1}{c|}{81.41} & 83.10 & 82.76 & 81.35 & 82.54 & \multicolumn{1}{c|}{81.36} & 83.52 & 82.65 & \multicolumn{1}{c|}{81.89} & 83.26 & 82.64 & 81.59 & 82.18 & \multicolumn{1}{c|}{80.16} & 83.13 & 82.09 & \multicolumn{1}{c|}{80.79} & 82.98 & 82.09 & 80.70 \\
\hline
\end{tabular}
}

\vspace{0.3cm}

\renewcommand\thetable{B.12}
\centering
\caption{Standard deviation comparison of test AUC scores (\%) on \texttt{blastchar} for various $(\alpha^{\text{tr}}, \alpha^{\text{ts}})$ combinations.}
\label{sup:tab:blastchar_std}
\fontsize{6pt}{7pt}\selectfont
\resizebox{0.99\textwidth}{!}{
\begin{tabular}{l|c|cccccccc|cccccccc|cccccccc}
\hline
\textbf{Method} & \textbf{No Missing} & \multicolumn{8}{c|}{\textbf{MAR}} & \multicolumn{8}{c|}{\textbf{MNAR}} & \multicolumn{8}{c}{\textbf{MCAR}} \\ \hline
$\alpha^{\text{tr}}$ & 0 & \multicolumn{2}{c|}{0} & \multicolumn{3}{c|}{0.15} & \multicolumn{3}{c|}{0.3} & \multicolumn{2}{c|}{0} & \multicolumn{3}{c|}{0.15} & \multicolumn{3}{c|}{0.3} & \multicolumn{2}{c|}{0} & \multicolumn{3}{c|}{0.15} & \multicolumn{3}{c}{0.3} \\ \hline
$\alpha^{\text{ts}}$ & 0 & 0.15 & \multicolumn{1}{c|}{0.3} & 0 & 0.15 & \multicolumn{1}{c|}{0.3} & 0 & 0.15 & 0.3 & 0.15 & \multicolumn{1}{c|}{0.3} & 0 & 0.15 & \multicolumn{1}{c|}{0.3} & 0 & 0.15 & 0.3 & 0.15 & \multicolumn{1}{c|}{0.3} & 0 & 0.15 & \multicolumn{1}{c|}{0.3} & 0 & 0.15 & 0.3 \\ \hline
LR-0 & 1.49 & 1.41 & \multicolumn{1}{c|}{1.42} & 0.92 & 1.03 & \multicolumn{1}{c|}{1.18} & 1.59 & 1.70 & 2.29 & 1.70 & \multicolumn{1}{c|}{1.38} & 1.48 & 1.18 & \multicolumn{1}{c|}{0.83} & 1.11 & 1.14 & 0.75 & 1.34 & \multicolumn{1}{c|}{1.89} & 2.90 & 3.15 & \multicolumn{1}{c|}{3.75} & 2.84 & 3.20 & 3.70 \\
NN-0 & 2.83 & 2.29 & \multicolumn{1}{c|}{1.94} & 1.37 & 1.76 & \multicolumn{1}{c|}{1.39} & 1.30 & 1.84 & 1.42 & 2.95 & \multicolumn{1}{c|}{3.12} & 1.05 & 0.94 & \multicolumn{1}{c|}{0.60} & 1.23 & 1.11 & 0.59 & 0.60 & \multicolumn{1}{c|}{0.70} & 0.93 & 1.46 & \multicolumn{1}{c|}{1.72} & 2.36 & 2.72 & 2.74 \\
RF & 0.56 & 0.53 & \multicolumn{1}{c|}{0.52} & 0.34 & 0.27 & \multicolumn{1}{c|}{0.24} & 0.52 & 0.36 & 0.17 & 1.12 & \multicolumn{1}{c|}{1.16} & 0.37 & 1.21 & \multicolumn{1}{c|}{1.37} & 0.55 & 0.99 & 1.48 & 0.74 & \multicolumn{1}{c|}{0.68} & 0.80 & 1.05 & \multicolumn{1}{c|}{0.71} & 0.68 & 0.88 & 0.87 \\
\hline
\end{tabular}
}

\vspace{0.3cm}

\renewcommand\thetable{B.13}
\centering
\caption{Comparison of averaged test AUC scores (\%) on \texttt{churn} for various $(\alpha^{\text{tr}},\alpha^{\text{ts}})$ combinations.}
\label{sup:tab:churn_auc}
\fontsize{6pt}{7pt}\selectfont
\resizebox{0.99\textwidth}{!}{
\begin{tabular}{l|c|cccccccc|cccccccc|cccccccc}
\hline
\textbf{Method} & \textbf{No Missing} & \multicolumn{8}{c|}{\textbf{MAR}} & \multicolumn{8}{c|}{\textbf{MNAR}} & \multicolumn{8}{c}{\textbf{MCAR}} \\ \hline
$\alpha^{\text{tr}}$ & 0 & \multicolumn{2}{c|}{0} & \multicolumn{3}{c|}{0.15} & \multicolumn{3}{c|}{0.3} & \multicolumn{2}{c|}{0} & \multicolumn{3}{c|}{0.15} & \multicolumn{3}{c|}{0.3} & \multicolumn{2}{c|}{0} & \multicolumn{3}{c|}{0.15} & \multicolumn{3}{c}{0.3} \\ \hline
$\alpha^{\text{ts}}$ & 0 & 0.15 & \multicolumn{1}{c|}{0.3} & 0 & 0.15 & \multicolumn{1}{c|}{0.3} & 0 & 0.15 & 0.3 & 0.15 & \multicolumn{1}{c|}{0.3} & 0 & 0.15 & \multicolumn{1}{c|}{0.3} & 0 & 0.15 & 0.3 & 0.15 & \multicolumn{1}{c|}{0.3} & 0 & 0.15 & \multicolumn{1}{c|}{0.3} & 0 & 0.15 & 0.3 \\ \hline
LR-0 & 53.92 & 51.25 & \multicolumn{1}{c|}{50.73} & 49.48 & 48.75 & \multicolumn{1}{c|}{48.59} & 47.46 & 47.63 & 47.65 & 52.54 & \multicolumn{1}{c|}{51.09} & 49.59 & 49.53 & \multicolumn{1}{c|}{48.81} & 47.01 & 47.73 & 47.45 & 51.71 & \multicolumn{1}{c|}{51.13} & 48.89 & 48.67 & \multicolumn{1}{c|}{48.58} & 47.22 & 47.50 & 47.49 \\
NN-0 & 58.11 & 56.64 & \multicolumn{1}{c|}{54.08} & 54.36 & 52.73 & \multicolumn{1}{c|}{52.77} & 53.85 & 53.81 & 54.44 & 57.85 & \multicolumn{1}{c|}{55.89} & 53.68 & 53.64 & \multicolumn{1}{c|}{54.32} & 52.59 & 53.37 & 54.50 & 57.13 & \multicolumn{1}{c|}{55.89} & 53.75 & 53.13 & \multicolumn{1}{c|}{53.33} & 53.47 & 53.30 & 54.24 \\
RF & 81.98 & 79.29 & \multicolumn{1}{c|}{75.33} & 82.35 & 79.28 & \multicolumn{1}{c|}{76.37} & 81.91 & 79.16 & 76.21 & 79.78 & \multicolumn{1}{c|}{75.18} & 82.70 & 80.24 & \multicolumn{1}{c|}{76.55} & 82.28 & 79.43 & 75.81 & 79.67 & \multicolumn{1}{c|}{75.35} & 82.35 & 79.69 & \multicolumn{1}{c|}{76.04} & 81.77 & 79.75 & 75.98 \\
\hline
\end{tabular}
}

\vspace{0.3cm}

\renewcommand\thetable{B.14}
\centering
\caption{Standard deviation comparison of test AUC scores (\%) on \texttt{churn} for various $(\alpha^{\text{tr}}, \alpha^{\text{ts}})$ combinations.}
\label{sup:tab:churn_std}
\fontsize{6pt}{7pt}\selectfont
\resizebox{0.99\textwidth}{!}{
\begin{tabular}{l|c|cccccccc|cccccccc|cccccccc}
\hline
\textbf{Method} & \textbf{No Missing} & \multicolumn{8}{c|}{\textbf{MAR}} & \multicolumn{8}{c|}{\textbf{MNAR}} & \multicolumn{8}{c}{\textbf{MCAR}} \\ \hline
$\alpha^{\text{tr}}$ & 0 & \multicolumn{2}{c|}{0} & \multicolumn{3}{c|}{0.15} & \multicolumn{3}{c|}{0.3} & \multicolumn{2}{c|}{0} & \multicolumn{3}{c|}{0.15} & \multicolumn{3}{c|}{0.3} & \multicolumn{2}{c|}{0} & \multicolumn{3}{c|}{0.15} & \multicolumn{3}{c}{0.3} \\ \hline
$\alpha^{\text{ts}}$ & 0 & 0.15 & \multicolumn{1}{c|}{0.3} & 0 & 0.15 & \multicolumn{1}{c|}{0.3} & 0 & 0.15 & 0.3 & 0.15 & \multicolumn{1}{c|}{0.3} & 0 & 0.15 & \multicolumn{1}{c|}{0.3} & 0 & 0.15 & 0.3 & 0.15 & \multicolumn{1}{c|}{0.3} & 0 & 0.15 & \multicolumn{1}{c|}{0.3} & 0 & 0.15 & 0.3 \\ \hline
LR-0 & 2.04 & 0.51 & \multicolumn{1}{c|}{0.31} & 1.37 & 0.61 & \multicolumn{1}{c|}{0.51} & 1.26 & 0.72 & 0.75 & 0.72 & \multicolumn{1}{c|}{0.40} & 2.75 & 1.53 & \multicolumn{1}{c|}{1.82} & 2.47 & 1.84 & 1.83 & 0.52 & \multicolumn{1}{c|}{0.57} & 1.47 & 1.06 & \multicolumn{1}{c|}{0.25} & 1.28 & 0.65 & 0.31 \\
NN-0 & 0.28 & 1.16 & \multicolumn{1}{c|}{0.25} & 4.16 & 2.15 & \multicolumn{1}{c|}{1.54} & 1.84 & 1.03 & 1.17 & 1.16 & \multicolumn{1}{c|}{0.43} & 2.68 & 2.56 & \multicolumn{1}{c|}{2.71} & 4.14 & 2.71 & 2.26 & 1.31 & \multicolumn{1}{c|}{0.29} & 1.70 & 1.73 & \multicolumn{1}{c|}{1.90} & 3.34 & 1.98 & 2.07 \\
RF & 2.14 & 1.22 & \multicolumn{1}{c|}{1.30} & 1.88 & 1.22 & \multicolumn{1}{c|}{0.88} & 1.27 & 0.85 & 0.79 & 1.30 & \multicolumn{1}{c|}{2.07} & 1.49 & 1.34 & \multicolumn{1}{c|}{1.98} & 1.56 & 1.05 & 1.59 & 1.95 & \multicolumn{1}{c|}{2.48} & 1.66 & 1.84 & \multicolumn{1}{c|}{1.21} & 1.64 & 1.66 & 0.95 \\
\hline
\end{tabular}
}
\end{sidewaystable}

\begin{sidewaystable}[htbp]
\renewcommand\thetable{B.15}
\centering
\caption{Comparison of averaged test AUC scores (\%) on \texttt{htru2} for various $(\alpha^{\text{tr}},\alpha^{\text{ts}})$ combinations.}
\label{sup:tab:htru2_auc}
\fontsize{6pt}{7pt}\selectfont
\resizebox{0.99\textwidth}{!}{
\begin{tabular}{l|c|cccccccc|cccccccc|cccccccc}
\hline
\textbf{Method} & \textbf{No Missing} & \multicolumn{8}{c|}{\textbf{MAR}} & \multicolumn{8}{c|}{\textbf{MNAR}} & \multicolumn{8}{c}{\textbf{MCAR}} \\ \hline
$\alpha^{\text{tr}}$ & 0 & \multicolumn{2}{c|}{0} & \multicolumn{3}{c|}{0.15} & \multicolumn{3}{c|}{0.3} & \multicolumn{2}{c|}{0} & \multicolumn{3}{c|}{0.15} & \multicolumn{3}{c|}{0.3} & \multicolumn{2}{c|}{0} & \multicolumn{3}{c|}{0.15} & \multicolumn{3}{c}{0.3} \\ \hline
$\alpha^{\text{ts}}$ & 0 & 0.15 & \multicolumn{1}{c|}{0.3} & 0 & 0.15 & \multicolumn{1}{c|}{0.3} & 0 & 0.15 & 0.3 & 0.15 & \multicolumn{1}{c|}{0.3} & 0 & 0.15 & \multicolumn{1}{c|}{0.3} & 0 & 0.15 & 0.3 & 0.15 & \multicolumn{1}{c|}{0.3} & 0 & 0.15 & \multicolumn{1}{c|}{0.3} & 0 & 0.15 & 0.3 \\ \hline
LR-0 & 96.65 & 89.66 & \multicolumn{1}{c|}{82.82} & 95.56 & 94.42 & \multicolumn{1}{c|}{92.48} & 95.40 & 94.29 & 92.53 & 89.45 & \multicolumn{1}{c|}{83.05} & 95.58 & 94.51 & \multicolumn{1}{c|}{92.71} & 95.50 & 94.42 & 92.64 & 90.00 & \multicolumn{1}{c|}{83.72} & 95.53 & 94.57 & \multicolumn{1}{c|}{92.45} & 95.42 & 94.51 & 92.55 \\
NN-0 & 96.89 & 90.69 & \multicolumn{1}{c|}{84.47} & 96.85 & 96.10 & \multicolumn{1}{c|}{95.11} & 96.66 & 96.19 & 95.51 & 91.17 & \multicolumn{1}{c|}{85.19} & 96.73 & 96.00 & \multicolumn{1}{c|}{94.81} & 96.58 & 96.00 & 95.09 & 91.31 & \multicolumn{1}{c|}{85.44} & 96.72 & 96.06 & \multicolumn{1}{c|}{94.78} & 96.62 & 96.16 & 95.25 \\
RF & 96.42 & 95.93 & \multicolumn{1}{c|}{92.79} & 96.68 & 96.50 & \multicolumn{1}{c|}{95.71} & 96.52 & 96.26 & 95.83 & 95.39 & \multicolumn{1}{c|}{95.04} & 96.63 & 96.24 & \multicolumn{1}{c|}{96.09} & 96.32 & 96.13 & 95.77 & 95.57 & \multicolumn{1}{c|}{94.03} & 96.32 & 96.34 & \multicolumn{1}{c|}{95.58} & 96.52 & 96.35 & 95.81 \\
\hline
\end{tabular}
}

\vspace{0.3cm}

\renewcommand\thetable{B.16}
\centering
\caption{Standard deviation comparison of test AUC scores (\%) on \texttt{htru2} for various $(\alpha^{\text{tr}}, \alpha^{\text{ts}})$ combinations.}
\label{sup:tab:htru2_std}
\fontsize{6pt}{7pt}\selectfont
\resizebox{0.99\textwidth}{!}{
\begin{tabular}{l|c|cccccccc|cccccccc|cccccccc}
\hline
\textbf{Method} & \textbf{No Missing} & \multicolumn{8}{c|}{\textbf{MAR}} & \multicolumn{8}{c|}{\textbf{MNAR}} & \multicolumn{8}{c}{\textbf{MCAR}} \\ \hline
$\alpha^{\text{tr}}$ & 0 & \multicolumn{2}{c|}{0} & \multicolumn{3}{c|}{0.15} & \multicolumn{3}{c|}{0.3} & \multicolumn{2}{c|}{0} & \multicolumn{3}{c|}{0.15} & \multicolumn{3}{c|}{0.3} & \multicolumn{2}{c|}{0} & \multicolumn{3}{c|}{0.15} & \multicolumn{3}{c}{0.3} \\ \hline
$\alpha^{\text{ts}}$ & 0 & 0.15 & \multicolumn{1}{c|}{0.3} & 0 & 0.15 & \multicolumn{1}{c|}{0.3} & 0 & 0.15 & 0.3 & 0.15 & \multicolumn{1}{c|}{0.3} & 0 & 0.15 & \multicolumn{1}{c|}{0.3} & 0 & 0.15 & 0.3 & 0.15 & \multicolumn{1}{c|}{0.3} & 0 & 0.15 & \multicolumn{1}{c|}{0.3} & 0 & 0.15 & 0.3 \\ \hline
LR-0 & 0.72 & 0.88 & \multicolumn{1}{c|}{1.85} & 0.62 & 0.70 & \multicolumn{1}{c|}{0.48} & 0.63 & 0.73 & 0.37 & 0.49 & \multicolumn{1}{c|}{0.39} & 0.58 & 0.68 & \multicolumn{1}{c|}{0.82} & 0.67 & 0.61 & 0.66 & 0.77 & \multicolumn{1}{c|}{1.04} & 0.58 & 0.70 & \multicolumn{1}{c|}{0.58} & 0.62 & 0.64 & 0.50 \\
NN-0 & 0.54 & 0.16 & \multicolumn{1}{c|}{0.45} & 0.44 & 0.35 & \multicolumn{1}{c|}{0.43} & 0.55 & 0.61 & 0.65 & 0.43 & \multicolumn{1}{c|}{1.40} & 0.46 & 0.40 & \multicolumn{1}{c|}{0.44} & 0.55 & 0.39 & 0.43 & 0.76 & \multicolumn{1}{c|}{1.48} & 0.45 & 0.35 & \multicolumn{1}{c|}{0.36} & 0.46 & 0.54 & 0.60 \\
RF & 0.88 & 0.80 & \multicolumn{1}{c|}{2.67} & 0.51 & 0.71 & \multicolumn{1}{c|}{0.65} & 0.48 & 0.52 & 0.80 & 0.49 & \multicolumn{1}{c|}{0.52} & 0.42 & 0.46 & \multicolumn{1}{c|}{0.33} & 0.72 & 0.49 & 0.48 & 0.42 & \multicolumn{1}{c|}{1.77} & 0.62 & 0.55 & \multicolumn{1}{c|}{0.31} & 0.57 & 0.40 & 0.25 \\
\hline
\end{tabular}
}

\vspace{0.3cm}

\renewcommand\thetable{B.17}
\centering
\caption{Comparison of averaged test AUC scores (\%) on \texttt{qsar\_bio} for various $(\alpha^{\text{tr}},\alpha^{\text{ts}})$ combinations.}
\label{sup:tab:qsar_bio_auc}
\fontsize{6pt}{7pt}\selectfont
\resizebox{0.99\textwidth}{!}{
\begin{tabular}{l|c|cccccccc|cccccccc|cccccccc}
\hline
\textbf{Method} & \textbf{No Missing} & \multicolumn{8}{c|}{\textbf{MAR}} & \multicolumn{8}{c|}{\textbf{MNAR}} & \multicolumn{8}{c}{\textbf{MCAR}} \\ \hline
$\alpha^{\text{tr}}$ & 0 & \multicolumn{2}{c|}{0} & \multicolumn{3}{c|}{0.15} & \multicolumn{3}{c|}{0.3} & \multicolumn{2}{c|}{0} & \multicolumn{3}{c|}{0.15} & \multicolumn{3}{c|}{0.3} & \multicolumn{2}{c|}{0} & \multicolumn{3}{c|}{0.15} & \multicolumn{3}{c}{0.3} \\ \hline
$\alpha^{\text{ts}}$ & 0 & 0.15 & \multicolumn{1}{c|}{0.3} & 0 & 0.15 & \multicolumn{1}{c|}{0.3} & 0 & 0.15 & 0.3 & 0.15 & \multicolumn{1}{c|}{0.3} & 0 & 0.15 & \multicolumn{1}{c|}{0.3} & 0 & 0.15 & 0.3 & 0.15 & \multicolumn{1}{c|}{0.3} & 0 & 0.15 & \multicolumn{1}{c|}{0.3} & 0 & 0.15 & 0.3 \\ \hline
LR-0 & 89.42 & 77.08 & \multicolumn{1}{c|}{70.38} & 87.68 & 79.06 & \multicolumn{1}{c|}{74.91} & 87.15 & 83.09 & 78.94 & 74.89 & \multicolumn{1}{c|}{72.01} & 88.10 & 81.91 & \multicolumn{1}{c|}{78.94} & 87.96 & 83.11 & 79.38 & 76.97 & \multicolumn{1}{c|}{70.91} & 87.87 & 81.91 & \multicolumn{1}{c|}{74.51} & 87.20 & 84.89 & 80.58 \\
NN-0 & 92.27 & 80.95 & \multicolumn{1}{c|}{72.63} & 91.63 & 87.45 & \multicolumn{1}{c|}{83.24} & 91.54 & 87.64 & 83.47 & 80.50 & \multicolumn{1}{c|}{76.92} & 91.21 & 87.21 & \multicolumn{1}{c|}{83.72} & 90.55 & 87.50 & 84.98 & 81.54 & \multicolumn{1}{c|}{75.48} & 91.45 & 87.19 & \multicolumn{1}{c|}{82.42} & 90.81 & 87.67 & 84.45 \\
RF & 86.10 & 84.50 & \multicolumn{1}{c|}{81.45} & 85.80 & 85.11 & \multicolumn{1}{c|}{82.35} & 84.34 & 83.35 & 81.26 & 84.28 & \multicolumn{1}{c|}{82.83} & 85.46 & 84.25 & \multicolumn{1}{c|}{83.02} & 84.96 & 83.46 & 82.98 & 84.82 & \multicolumn{1}{c|}{84.08} & 86.93 & 85.70 & \multicolumn{1}{c|}{83.86} & 86.66 & 85.77 & 83.90 \\
\hline
\end{tabular}
}

\vspace{0.3cm}

\renewcommand\thetable{B.18}
\centering
\caption{Standard deviation comparison of test AUC scores (\%) on \texttt{qsar\_bio} for various $(\alpha^{\text{tr}}, \alpha^{\text{ts}})$ combinations.}
\label{sup:tab:qsar_bio_std}
\fontsize{6pt}{7pt}\selectfont
\resizebox{0.99\textwidth}{!}{
\begin{tabular}{l|c|cccccccc|cccccccc|cccccccc}
\hline
\textbf{Method} & \textbf{No Missing} & \multicolumn{8}{c|}{\textbf{MAR}} & \multicolumn{8}{c|}{\textbf{MNAR}} & \multicolumn{8}{c}{\textbf{MCAR}} \\ \hline
$\alpha^{\text{tr}}$ & 0 & \multicolumn{2}{c|}{0} & \multicolumn{3}{c|}{0.15} & \multicolumn{3}{c|}{0.3} & \multicolumn{2}{c|}{0} & \multicolumn{3}{c|}{0.15} & \multicolumn{3}{c|}{0.3} & \multicolumn{2}{c|}{0} & \multicolumn{3}{c|}{0.15} & \multicolumn{3}{c}{0.3} \\ \hline
$\alpha^{\text{ts}}$ & 0 & 0.15 & \multicolumn{1}{c|}{0.3} & 0 & 0.15 & \multicolumn{1}{c|}{0.3} & 0 & 0.15 & 0.3 & 0.15 & \multicolumn{1}{c|}{0.3} & 0 & 0.15 & \multicolumn{1}{c|}{0.3} & 0 & 0.15 & 0.3 & 0.15 & \multicolumn{1}{c|}{0.3} & 0 & 0.15 & \multicolumn{1}{c|}{0.3} & 0 & 0.15 & 0.3 \\ \hline
LR-0 & 2.22 & 2.00 & \multicolumn{1}{c|}{3.72} & 0.54 & 1.22 & \multicolumn{1}{c|}{1.08} & 2.19 & 1.85 & 0.93 & 2.32 & \multicolumn{1}{c|}{3.21} & 1.40 & 1.10 & \multicolumn{1}{c|}{1.85} & 1.55 & 1.10 & 1.99 & 1.68 & \multicolumn{1}{c|}{3.28} & 1.41 & 2.45 & \multicolumn{1}{c|}{4.78} & 1.25 & 1.67 & 2.59 \\
NN-0 & 1.20 & 2.30 & \multicolumn{1}{c|}{4.73} & 0.29 & 0.93 & \multicolumn{1}{c|}{1.14} & 0.93 & 1.35 & 2.85 & 1.37 & \multicolumn{1}{c|}{1.91} & 1.19 & 0.99 & \multicolumn{1}{c|}{1.72} & 0.74 & 0.20 & 1.11 & 1.57 & \multicolumn{1}{c|}{3.23} & 0.19 & 1.52 & \multicolumn{1}{c|}{2.32} & 0.08 & 1.35 & 0.65 \\
RF & 0.19 & 1.52 & \multicolumn{1}{c|}{0.40} & 0.35 & 0.32 & \multicolumn{1}{c|}{1.29} & 0.42 & 1.57 & 0.75 & 0.69 & \multicolumn{1}{c|}{1.32} & 0.99 & 1.15 & \multicolumn{1}{c|}{0.91} & 1.14 & 0.47 & 2.11 & 0.97 & \multicolumn{1}{c|}{0.36} & 1.36 & 1.53 & \multicolumn{1}{c|}{1.92} & 0.82 & 1.40 & 1.72 \\
\hline
\end{tabular}
}
\end{sidewaystable}

\begin{sidewaystable}[htbp]
\renewcommand\thetable{B.19}
\centering
\caption{Comparison of averaged test AUC scores (\%) on \texttt{shoppers} for various $(\alpha^{\text{tr}},\alpha^{\text{ts}})$ combinations.}
\label{sup:tab:shoppers_auc}
\fontsize{6pt}{7pt}\selectfont
\resizebox{0.99\textwidth}{!}{
\begin{tabular}{l|c|cccccccc|cccccccc|cccccccc}
\hline
\textbf{Method} & \textbf{No Missing} & \multicolumn{8}{c|}{\textbf{MAR}} & \multicolumn{8}{c|}{\textbf{MNAR}} & \multicolumn{8}{c}{\textbf{MCAR}} \\ \hline
$\alpha^{\text{tr}}$ & 0 & \multicolumn{2}{c|}{0} & \multicolumn{3}{c|}{0.15} & \multicolumn{3}{c|}{0.3} & \multicolumn{2}{c|}{0} & \multicolumn{3}{c|}{0.15} & \multicolumn{3}{c|}{0.3} & \multicolumn{2}{c|}{0} & \multicolumn{3}{c|}{0.15} & \multicolumn{3}{c}{0.3} \\ \hline
$\alpha^{\text{ts}}$ & 0 & 0.15 & \multicolumn{1}{c|}{0.3} & 0 & 0.15 & \multicolumn{1}{c|}{0.3} & 0 & 0.15 & 0.3 & 0.15 & \multicolumn{1}{c|}{0.3} & 0 & 0.15 & \multicolumn{1}{c|}{0.3} & 0 & 0.15 & 0.3 & 0.15 & \multicolumn{1}{c|}{0.3} & 0 & 0.15 & \multicolumn{1}{c|}{0.3} & 0 & 0.15 & 0.3 \\ \hline
LR-0 & 57.00 & 55.28 & \multicolumn{1}{c|}{53.20} & 58.39 & 56.45 & \multicolumn{1}{c|}{54.07} & 57.46 & 55.93 & 53.72 & 55.80 & \multicolumn{1}{c|}{54.31} & 55.71 & 54.60 & \multicolumn{1}{c|}{53.30} & 53.14 & 52.51 & 51.57 & 55.67 & \multicolumn{1}{c|}{53.88} & 53.38 & 52.91 & \multicolumn{1}{c|}{51.83} & 52.42 & 51.46 & 50.51 \\
NN-0 & 91.47 & 86.93 & \multicolumn{1}{c|}{82.60} & 90.85 & 87.35 & \multicolumn{1}{c|}{83.59} & 90.50 & 87.45 & 83.85 & 86.65 & \multicolumn{1}{c|}{81.59} & 91.14 & 87.34 & \multicolumn{1}{c|}{83.38} & 90.75 & 87.12 & 83.45 & 86.49 & \multicolumn{1}{c|}{81.52} & 91.19 & 87.36 & \multicolumn{1}{c|}{83.05} & 91.25 & 87.73 & 83.48 \\
RF & 91.59 & 87.02 & \multicolumn{1}{c|}{82.18} & 90.90 & 88.77 & \multicolumn{1}{c|}{85.83} & 90.00 & 87.35 & 85.30 & 85.74 & \multicolumn{1}{c|}{79.66} & 90.54 & 88.75 & \multicolumn{1}{c|}{85.93} & 90.05 & 88.41 & 85.63 & 86.58 & \multicolumn{1}{c|}{80.88} & 90.86 & 88.79 & \multicolumn{1}{c|}{85.62} & 90.06 & 88.38 & 85.59 \\
\hline
\end{tabular}
}

\vspace{0.3cm}

\renewcommand\thetable{B.20}
\centering
\caption{Standard deviation comparison of test AUC scores (\%) on \texttt{shoppers} for various $(\alpha^{\text{tr}}, \alpha^{\text{ts}})$ combinations.}
\label{sup:tab:shoppers_std}
\fontsize{6pt}{7pt}\selectfont
\resizebox{0.99\textwidth}{!}{
\begin{tabular}{l|c|cccccccc|cccccccc|cccccccc}
\hline
\textbf{Method} & \textbf{No Missing} & \multicolumn{8}{c|}{\textbf{MAR}} & \multicolumn{8}{c|}{\textbf{MNAR}} & \multicolumn{8}{c}{\textbf{MCAR}} \\ \hline
$\alpha^{\text{tr}}$ & 0 & \multicolumn{2}{c|}{0} & \multicolumn{3}{c|}{0.15} & \multicolumn{3}{c|}{0.3} & \multicolumn{2}{c|}{0} & \multicolumn{3}{c|}{0.15} & \multicolumn{3}{c|}{0.3} & \multicolumn{2}{c|}{0} & \multicolumn{3}{c|}{0.15} & \multicolumn{3}{c}{0.3} \\ \hline
$\alpha^{\text{ts}}$ & 0 & 0.15 & \multicolumn{1}{c|}{0.3} & 0 & 0.15 & \multicolumn{1}{c|}{0.3} & 0 & 0.15 & 0.3 & 0.15 & \multicolumn{1}{c|}{0.3} & 0 & 0.15 & \multicolumn{1}{c|}{0.3} & 0 & 0.15 & 0.3 & 0.15 & \multicolumn{1}{c|}{0.3} & 0 & 0.15 & \multicolumn{1}{c|}{0.3} & 0 & 0.15 & 0.3 \\ \hline
LR-0 & 2.24 & 2.27 & \multicolumn{1}{c|}{2.19} & 1.66 & 1.70 & \multicolumn{1}{c|}{1.36} & 2.08 & 2.25 & 1.92 & 1.62 & \multicolumn{1}{c|}{0.75} & 3.01 & 2.80 & \multicolumn{1}{c|}{1.86} & 3.46 & 3.05 & 2.19 & 0.73 & \multicolumn{1}{c|}{0.37} & 1.65 & 0.45 & \multicolumn{1}{c|}{0.82} & 1.43 & 0.60 & 0.09 \\
NN-0 & 0.42 & 0.48 & \multicolumn{1}{c|}{0.84} & 0.74 & 0.53 & \multicolumn{1}{c|}{0.35} & 0.50 & 0.38 & 0.35 & 0.51 & \multicolumn{1}{c|}{0.46} & 0.39 & 0.18 & \multicolumn{1}{c|}{0.26} & 0.57 & 0.48 & 0.16 & 1.19 & \multicolumn{1}{c|}{1.05} & 0.08 & 0.20 & \multicolumn{1}{c|}{0.46} & 0.36 & 0.48 & 0.32 \\
RF & 0.42 & 0.40 & \multicolumn{1}{c|}{0.60} & 0.52 & 0.35 & \multicolumn{1}{c|}{0.61} & 0.60 & 0.48 & 0.53 & 0.08 & \multicolumn{1}{c|}{1.79} & 0.21 & 0.37 & \multicolumn{1}{c|}{0.87} & 0.64 & 0.50 & 0.72 & 0.44 & \multicolumn{1}{c|}{0.57} & 0.27 & 0.30 & \multicolumn{1}{c|}{0.42} & 0.17 & 0.16 & 0.30 \\
\hline
\end{tabular}
}

\vspace{0.3cm}

\renewcommand\thetable{B.21}
\centering
\caption{Comparison of averaged test AUC scores (\%) on \texttt{spambase} for various $(\alpha^{\text{tr}},\alpha^{\text{ts}})$ combinations.}
\label{sup:tab:spambase_auc}
\fontsize{6pt}{7pt}\selectfont
\resizebox{0.99\textwidth}{!}{
\begin{tabular}{l|c|cccccccc|cccccccc|cccccccc}
\hline
\textbf{Method} & \textbf{No Missing} & \multicolumn{8}{c|}{\textbf{MAR}} & \multicolumn{8}{c|}{\textbf{MNAR}} & \multicolumn{8}{c}{\textbf{MCAR}} \\ \hline
$\alpha^{\text{tr}}$ & 0 & \multicolumn{2}{c|}{0} & \multicolumn{3}{c|}{0.15} & \multicolumn{3}{c|}{0.3} & \multicolumn{2}{c|}{0} & \multicolumn{3}{c|}{0.15} & \multicolumn{3}{c|}{0.3} & \multicolumn{2}{c|}{0} & \multicolumn{3}{c|}{0.15} & \multicolumn{3}{c}{0.3} \\ \hline
$\alpha^{\text{ts}}$ & 0 & 0.15 & \multicolumn{1}{c|}{0.3} & 0 & 0.15 & \multicolumn{1}{c|}{0.3} & 0 & 0.15 & 0.3 & 0.15 & \multicolumn{1}{c|}{0.3} & 0 & 0.15 & \multicolumn{1}{c|}{0.3} & 0 & 0.15 & 0.3 & 0.15 & \multicolumn{1}{c|}{0.3} & 0 & 0.15 & \multicolumn{1}{c|}{0.3} & 0 & 0.15 & 0.3 \\ \hline
LR-0 & 84.44 & 77.81 & \multicolumn{1}{c|}{72.04} & 84.56 & 79.99 & \multicolumn{1}{c|}{75.62} & 84.33 & 81.29 & 77.96 & 77.92 & \multicolumn{1}{c|}{71.77} & 84.19 & 79.15 & \multicolumn{1}{c|}{74.73} & 83.53 & 79.53 & 75.38 & 77.38 & \multicolumn{1}{c|}{71.28} & 84.52 & 79.51 & \multicolumn{1}{c|}{74.88} & 84.10 & 81.16 & 77.26 \\
NN-0 & 97.75 & 92.19 & \multicolumn{1}{c|}{85.72} & 97.69 & 96.89 & \multicolumn{1}{c|}{95.90} & 97.59 & 96.87 & 96.00 & 92.73 & \multicolumn{1}{c|}{87.92} & 97.69 & 97.00 & \multicolumn{1}{c|}{95.82} & 97.45 & 96.91 & 95.94 & 92.09 & \multicolumn{1}{c|}{86.47} & 97.61 & 96.73 & \multicolumn{1}{c|}{95.69} & 97.47 & 96.77 & 95.92 \\
RF & 96.47 & 94.31 & \multicolumn{1}{c|}{92.89} & 95.34 & 94.44 & \multicolumn{1}{c|}{93.75} & 94.88 & 94.12 & 93.50 & 94.65 & \multicolumn{1}{c|}{93.07} & 95.51 & 95.15 & \multicolumn{1}{c|}{94.17} & 95.09 & 94.61 & 93.79 & 94.38 & \multicolumn{1}{c|}{92.94} & 95.39 & 94.62 & \multicolumn{1}{c|}{93.71} & 94.82 & 94.15 & 93.38 \\
\hline
\end{tabular}
}

\vspace{0.3cm}

\renewcommand\thetable{B.22}
\centering
\caption{Standard deviation comparison of test AUC scores (\%) on \texttt{spambase} for various $(\alpha^{\text{tr}}, \alpha^{\text{ts}})$ combinations.}
\label{sup:tab:spambase_std}
\fontsize{6pt}{7pt}\selectfont
\resizebox{0.99\textwidth}{!}{
\begin{tabular}{l|c|cccccccc|cccccccc|cccccccc}
\hline
\textbf{Method} & \textbf{No Missing} & \multicolumn{8}{c|}{\textbf{MAR}} & \multicolumn{8}{c|}{\textbf{MNAR}} & \multicolumn{8}{c}{\textbf{MCAR}} \\ \hline
$\alpha^{\text{tr}}$ & 0 & \multicolumn{2}{c|}{0} & \multicolumn{3}{c|}{0.15} & \multicolumn{3}{c|}{0.3} & \multicolumn{2}{c|}{0} & \multicolumn{3}{c|}{0.15} & \multicolumn{3}{c|}{0.3} & \multicolumn{2}{c|}{0} & \multicolumn{3}{c|}{0.15} & \multicolumn{3}{c}{0.3} \\ \hline
$\alpha^{\text{ts}}$ & 0 & 0.15 & \multicolumn{1}{c|}{0.3} & 0 & 0.15 & \multicolumn{1}{c|}{0.3} & 0 & 0.15 & 0.3 & 0.15 & \multicolumn{1}{c|}{0.3} & 0 & 0.15 & \multicolumn{1}{c|}{0.3} & 0 & 0.15 & 0.3 & 0.15 & \multicolumn{1}{c|}{0.3} & 0 & 0.15 & \multicolumn{1}{c|}{0.3} & 0 & 0.15 & 0.3 \\ \hline
LR-0 & 3.96 & 3.12 & \multicolumn{1}{c|}{3.75} & 3.33 & 3.84 & \multicolumn{1}{c|}{4.58} & 3.25 & 4.19 & 5.34 & 4.09 & \multicolumn{1}{c|}{5.69} & 3.54 & 5.79 & \multicolumn{1}{c|}{8.58} & 2.95 & 4.63 & 7.41 & 4.56 & \multicolumn{1}{c|}{5.43} & 3.69 & 5.64 & \multicolumn{1}{c|}{6.07} & 3.16 & 4.59 & 5.71 \\
NN-0 & 0.60 & 0.85 & \multicolumn{1}{c|}{1.03} & 0.37 & 0.11 & \multicolumn{1}{c|}{0.28} & 0.57 & 0.21 & 0.25 & 1.88 & \multicolumn{1}{c|}{2.34} & 0.43 & 0.56 & \multicolumn{1}{c|}{0.81} & 0.43 & 0.65 & 0.87 & 1.46 & \multicolumn{1}{c|}{2.79} & 0.45 & 0.51 & \multicolumn{1}{c|}{0.43} & 0.51 & 0.58 & 0.42 \\
RF & 0.83 & 0.63 & \multicolumn{1}{c|}{1.41} & 0.47 & 0.20 & \multicolumn{1}{c|}{0.59} & 0.33 & 0.33 & 0.20 & 0.73 & \multicolumn{1}{c|}{1.06} & 0.22 & 0.43 & \multicolumn{1}{c|}{0.40} & 0.54 & 0.70 & 0.31 & 0.46 & \multicolumn{1}{c|}{0.61} & 0.48 & 0.68 & \multicolumn{1}{c|}{0.47} & 0.46 & 0.49 & 0.36 \\
\hline
\end{tabular}
}
\end{sidewaystable}

\newpage



\subsection{B-\MakeUppercase{\romannumeral 5}. Detailed Results for Input-Output-Missingness Scenarios}
\label{sec:appB5}
Tables~\ref{ssl:tab:adult}-\ref{ssl:tab:spambase} report the average AUC scores, obtained from three independent runs of each method across all benchmark datasets under the input-output missingness settings.

\begin{table}[H]
\renewcommand\thetable{B.23}
\centering
\fontsize{6pt}{7pt}\selectfont
\caption{Comparison of averaged test AUC scores (\%) on \texttt{adult} for various $(\alpha^{\text{tr}},\alpha^{\text{ts}},\alpha_y^{\text{tr}})$ combinations.}
\resizebox{0.99\textwidth}{!}{
\begin{tabular}{cl|llllll|llllll|llllll}
\toprule
\multicolumn{2}{c|}{\textbf{Method}} & \multicolumn{6}{c|}{\textbf{MAR}} & \multicolumn{6}{c|}{\textbf{MNAR}} & \multicolumn{6}{c}{\textbf{MCAR}} \\ \midrule
\multicolumn{1}{c|}{\multirow{2}{*}{$\alpha_y^{\text{tr}}$}} & \textbf{$\alpha^{\text{tr}}$} & \multicolumn{3}{c|}{\textbf{0.15}} & \multicolumn{3}{c|}{\textbf{0.3}} & \multicolumn{3}{c|}{\textbf{0.15}} & \multicolumn{3}{c|}{\textbf{0.3}} & \multicolumn{3}{c|}{\textbf{0.15}} & \multicolumn{3}{c}{\textbf{0.3}} \\ \cline{2-20} 
\multicolumn{1}{c|}{} & \textbf{$\alpha^{\text{ts}}$} & \multicolumn{1}{c}{\textbf{0}} & \multicolumn{1}{c}{\textbf{0.15}} & \multicolumn{1}{c|}{\textbf{0.3}} & \multicolumn{1}{c}{\textbf{0}} & \multicolumn{1}{c}{\textbf{0.15}} & \multicolumn{1}{c|}{\textbf{0.3}} & \multicolumn{1}{c}{\textbf{0}} & \multicolumn{1}{c}{\textbf{0.15}} & \multicolumn{1}{c|}{\textbf{0.3}} & \multicolumn{1}{c}{\textbf{0}} & \multicolumn{1}{c}{\textbf{0.15}} & \multicolumn{1}{c|}{\textbf{0.3}} & \multicolumn{1}{c}{\textbf{0}} & \multicolumn{1}{c}{\textbf{0.15}} & \multicolumn{1}{c|}{\textbf{0.3}} & \multicolumn{1}{c}{\textbf{0}} & \multicolumn{1}{c}{\textbf{0.15}} & \multicolumn{1}{c}{\textbf{0.3}} \\ \midrule
\multicolumn{1}{c|}{\multirow{3}{*}{0.15}} & SAINT & 91.52 & 90.61 & \multicolumn{1}{l|}{89.24} & 91.26 & 90.40 & 89.20 & 91.47 & 90.59 & \multicolumn{1}{l|}{88.96} & 91.27 & 90.47 & 89.11 & 91.59 & 90.72 & \multicolumn{1}{l|}{89.16} & 91.33 & 90.56 & 89.28 \\
\multicolumn{1}{c|}{} & Stab & 89.95 & 88.67 & \multicolumn{1}{l|}{86.86} & 89.28 & 88.21 & 86.55 & 89.82 & 88.59 & \multicolumn{1}{l|}{86.79} & 89.57 & 88.59 & 86.83 & 89.86 & 88.84 & \multicolumn{1}{l|}{86.98} & 89.60 & 88.71 & 87.13 \\ \cline{2-20} 
\multicolumn{1}{c|}{} & Ours & 91.20 & 90.20 & \multicolumn{1}{l|}{88.83} & 91.22 & 90.32 & 89.07 & 91.21 & 90.38 & \multicolumn{1}{l|}{88.86} & 91.14 & 90.29 & 88.82 & 91.25 & 90.43 & \multicolumn{1}{l|}{88.99} & 91.13 & 90.35 & 88.91 \\
\bottomrule
\end{tabular}
}
\label{ssl:tab:adult}
\vspace{0.4cm}

\renewcommand\thetable{B.24}
\centering
\fontsize{6pt}{7pt}\selectfont
\caption{Comparison of averaged test AUC scores (\%) on \texttt{arcene} for various $(\alpha^{\text{tr}},\alpha^{\text{ts}},\alpha_y^{\text{tr}})$ combinations.}
\resizebox{0.99\textwidth}{!}{
\begin{tabular}{cl|llllll|llllll|llllll}
\toprule
\multicolumn{2}{c|}{\textbf{Method}} & \multicolumn{6}{c|}{\textbf{MAR}} & \multicolumn{6}{c|}{\textbf{MNAR}} & \multicolumn{6}{c}{\textbf{MCAR}} \\ \midrule
\multicolumn{1}{c|}{\multirow{2}{*}{$\alpha_y^{\text{tr}}$}} & \textbf{$\alpha^{\text{tr}}$} & \multicolumn{3}{c|}{\textbf{0.15}} & \multicolumn{3}{c|}{\textbf{0.3}} & \multicolumn{3}{c|}{\textbf{0.15}} & \multicolumn{3}{c|}{\textbf{0.3}} & \multicolumn{3}{c|}{\textbf{0.15}} & \multicolumn{3}{c}{\textbf{0.3}} \\ \cline{2-20} 
\multicolumn{1}{c|}{} & \textbf{$\alpha^{\text{ts}}$} & \multicolumn{1}{c}{\textbf{0}} & \multicolumn{1}{c}{\textbf{0.15}} & \multicolumn{1}{c|}{\textbf{0.3}} & \multicolumn{1}{c}{\textbf{0}} & \multicolumn{1}{c}{\textbf{0.15}} & \multicolumn{1}{c|}{\textbf{0.3}} & \multicolumn{1}{c}{\textbf{0}} & \multicolumn{1}{c}{\textbf{0.15}} & \multicolumn{1}{c|}{\textbf{0.3}} & \multicolumn{1}{c}{\textbf{0}} & \multicolumn{1}{c}{\textbf{0.15}} & \multicolumn{1}{c|}{\textbf{0.3}} & \multicolumn{1}{c}{\textbf{0}} & \multicolumn{1}{c}{\textbf{0.15}} & \multicolumn{1}{c|}{\textbf{0.3}} & \multicolumn{1}{c}{\textbf{0}} & \multicolumn{1}{c}{\textbf{0.15}} & \multicolumn{1}{c}{\textbf{0.3}} \\ \midrule
\multicolumn{1}{c|}{\multirow{3}{*}{0.15}} & SAINT & 74.42 & 74.42 & \multicolumn{1}{l|}{73.61} & 72.34 & 74.65 & 71.30 & 73.15 & 74.07 & \multicolumn{1}{l|}{73.96} & 69.91 & 71.30 & 70.72 & 73.61 & 73.73 & \multicolumn{1}{l|}{74.77} & 71.88 & 71.41 & 72.11 \\
\multicolumn{1}{c|}{} & Stab & 89.47 & 85.19 & \multicolumn{1}{l|}{77.55} & 85.42 & 81.13 & 79.28 & 84.95 & 87.62 & \multicolumn{1}{l|}{82.06} & 89.24 & 89.93 & 87.27 & 91.32 & 86.69 & \multicolumn{1}{l|}{78.70} & 89.35 & 86.00 & 80.79 \\ \cline{2-20} 
\multicolumn{1}{c|}{} & Ours & 91.55 & 90.97 & \multicolumn{1}{l|}{89.41} & 87.96 & 87.79 & 87.85 & 90.91 & 90.91 & \multicolumn{1}{l|}{91.61} & 89.06 & 90.39 & 91.26 & 91.26 & 89.93 & \multicolumn{1}{l|}{89.47} & 87.62 & 86.34 & 86.69 \\
\bottomrule
\end{tabular}
}
\label{ssl:tab:arcene}
\vspace{0.4cm}

\renewcommand\thetable{B.25}
\centering
\fontsize{6pt}{7pt}\selectfont
\caption{Comparison of averaged test AUC scores (\%) on \texttt{arrhythmia} for various $(\alpha^{\text{tr}},\alpha^{\text{ts}},\alpha_y^{\text{tr}})$ combinations.}
\resizebox{0.99\textwidth}{!}{
\begin{tabular}{cl|llllll|llllll|llllll}
\toprule
\multicolumn{2}{c|}{\textbf{Method}} & \multicolumn{6}{c|}{\textbf{MAR}} & \multicolumn{6}{c|}{\textbf{MNAR}} & \multicolumn{6}{c}{\textbf{MCAR}} \\ \midrule
\multicolumn{1}{c|}{\multirow{2}{*}{$\alpha_y^{\text{tr}}$}} & \textbf{$\alpha^{\text{tr}}$} & \multicolumn{3}{c|}{\textbf{0.15}} & \multicolumn{3}{c|}{\textbf{0.3}} & \multicolumn{3}{c|}{\textbf{0.15}} & \multicolumn{3}{c|}{\textbf{0.3}} & \multicolumn{3}{c|}{\textbf{0.15}} & \multicolumn{3}{c}{\textbf{0.3}} \\ \cline{2-20} 
\multicolumn{1}{c|}{} & \textbf{$\alpha^{\text{ts}}$} & \multicolumn{1}{c}{\textbf{0}} & \multicolumn{1}{c}{\textbf{0.15}} & \multicolumn{1}{c|}{\textbf{0.3}} & \multicolumn{1}{c}{\textbf{0}} & \multicolumn{1}{c}{\textbf{0.15}} & \multicolumn{1}{c|}{\textbf{0.3}} & \multicolumn{1}{c}{\textbf{0}} & \multicolumn{1}{c}{\textbf{0.15}} & \multicolumn{1}{c|}{\textbf{0.3}} & \multicolumn{1}{c}{\textbf{0}} & \multicolumn{1}{c}{\textbf{0.15}} & \multicolumn{1}{c|}{\textbf{0.3}} & \multicolumn{1}{c}{\textbf{0}} & \multicolumn{1}{c}{\textbf{0.15}} & \multicolumn{1}{c|}{\textbf{0.3}} & \multicolumn{1}{c}{\textbf{0}} & \multicolumn{1}{c}{\textbf{0.15}} & \multicolumn{1}{c}{\textbf{0.3}} \\ \midrule
\multicolumn{1}{c|}{\multirow{3}{*}{0.15}} & SAINT & 82.91 & 81.48 & \multicolumn{1}{l|}{79.24} & 82.70 & 82.83 & 82.45 & 82.15 & 78.35 & \multicolumn{1}{l|}{77.51} & 80.51 & 76.88 & 75.23 & 81.77 & 81.14 & \multicolumn{1}{l|}{80.59} & 67.38 & 62.32 & 60.04 \\
\multicolumn{1}{c|}{} & Stab & 61.25 & 57.68 & \multicolumn{1}{l|}{55.74} & 61.29 & 59.28 & 56.84 & 67.93 & 55.27 & \multicolumn{1}{l|}{46.52} & 69.14 & 54.05 & 47.26 & 66.96 & 62.11 & \multicolumn{1}{l|}{58.73} & 66.14 & 58.50 & 56.18 \\ \cline{2-20} 
\multicolumn{1}{c|}{} & Ours & 79.54 & 75.70 & \multicolumn{1}{l|}{74.30} & 80.72 & 75.40 & 77.64 & 83.50 & 81.90 & \multicolumn{1}{l|}{81.31} & 81.94 & 82.03 & 80.04 & 79.45 & 78.06 & \multicolumn{1}{l|}{80.80} & 79.28 & 81.65 & 78.95 \\
\bottomrule
\end{tabular}
}
\label{ssl:tab:arrhythmia}
\vspace{0.4cm}

\renewcommand\thetable{B.26}
\centering
\fontsize{6pt}{7pt}\selectfont
\caption{Comparison of averaged test AUC scores (\%) on \texttt{bank} for various $(\alpha^{\text{tr}},\alpha^{\text{ts}},\alpha_y^{\text{tr}})$ combinations.}
\resizebox{0.99\textwidth}{!}{
\begin{tabular}{cl|llllll|llllll|llllll}
\toprule
\multicolumn{2}{c|}{\textbf{Method}} & \multicolumn{6}{c|}{\textbf{MAR}} & \multicolumn{6}{c|}{\textbf{MNAR}} & \multicolumn{6}{c}{\textbf{MCAR}} \\ \midrule
\multicolumn{1}{c|}{\multirow{2}{*}{$\alpha_y^{\text{tr}}$}} & \textbf{$\alpha^{\text{tr}}$} & \multicolumn{3}{c|}{\textbf{0.15}} & \multicolumn{3}{c|}{\textbf{0.3}} & \multicolumn{3}{c|}{\textbf{0.15}} & \multicolumn{3}{c|}{\textbf{0.3}} & \multicolumn{3}{c|}{\textbf{0.15}} & \multicolumn{3}{c}{\textbf{0.3}} \\ \cline{2-20} 
\multicolumn{1}{c|}{} & \textbf{$\alpha^{\text{ts}}$} & \multicolumn{1}{c}{\textbf{0}} & \multicolumn{1}{c}{\textbf{0.15}} & \multicolumn{1}{c|}{\textbf{0.3}} & \multicolumn{1}{c}{\textbf{0}} & \multicolumn{1}{c}{\textbf{0.15}} & \multicolumn{1}{c|}{\textbf{0.3}} & \multicolumn{1}{c}{\textbf{0}} & \multicolumn{1}{c}{\textbf{0.15}} & \multicolumn{1}{c|}{\textbf{0.3}} & \multicolumn{1}{c}{\textbf{0}} & \multicolumn{1}{c}{\textbf{0.15}} & \multicolumn{1}{c|}{\textbf{0.3}} & \multicolumn{1}{c}{\textbf{0}} & \multicolumn{1}{c}{\textbf{0.15}} & \multicolumn{1}{c|}{\textbf{0.3}} & \multicolumn{1}{c}{\textbf{0}} & \multicolumn{1}{c}{\textbf{0.15}} & \multicolumn{1}{c}{\textbf{0.3}} \\ \midrule
\multicolumn{1}{c|}{\multirow{3}{*}{0.15}} & SAINT & 92.58 & 90.30 & \multicolumn{1}{l|}{86.95} & 91.99 & 89.91 & 86.99 & 92.61 & 90.07 & \multicolumn{1}{l|}{87.38} & 91.89 & 89.81 & 87.29 & 92.61 & 90.30 & \multicolumn{1}{l|}{86.92} & 92.11 & 89.96 & 87.01 \\
\multicolumn{1}{c|}{} & Stab & 90.55 & 87.07 & \multicolumn{1}{l|}{83.61} & 89.39 & 86.56 & 83.48 & 90.49 & 87.20 & \multicolumn{1}{l|}{83.83} & 89.85 & 86.88 & 83.71 & 90.36 & 87.52 & \multicolumn{1}{l|}{83.98} & 89.45 & 86.91 & 83.61 \\ \cline{2-20} 
\multicolumn{1}{c|}{} & Ours & 90.61 & 88.21 & \multicolumn{1}{l|}{85.27} & 90.58 & 88.27 & 85.41 & 91.19 & 88.87 & \multicolumn{1}{l|}{86.36} & 90.73 & 88.39 & 85.81 & 90.67 & 88.42 & \multicolumn{1}{l|}{85.40} & 90.72 & 88.49 & 85.69 \\
\bottomrule
\end{tabular}
}
\label{ssl:tab:bank}
\vspace{0.4cm}

\renewcommand\thetable{B.27}
\centering
\fontsize{6pt}{7pt}\selectfont
\caption{Comparison of averaged test AUC scores (\%) on \texttt{blastchar} for various $(\alpha^{\text{tr}},\alpha^{\text{ts}},\alpha_y^{\text{tr}})$ combinations.}
\resizebox{0.99\textwidth}{!}{
\begin{tabular}{cl|llllll|llllll|llllll}
\toprule
\multicolumn{2}{c|}{\textbf{Method}} & \multicolumn{6}{c|}{\textbf{MAR}} & \multicolumn{6}{c|}{\textbf{MNAR}} & \multicolumn{6}{c}{\textbf{MCAR}} \\ \midrule
\multicolumn{1}{c|}{\multirow{2}{*}{$\alpha_y^{\text{tr}}$}} & \textbf{$\alpha^{\text{tr}}$} & \multicolumn{3}{c|}{\textbf{0.15}} & \multicolumn{3}{c|}{\textbf{0.3}} & \multicolumn{3}{c|}{\textbf{0.15}} & \multicolumn{3}{c|}{\textbf{0.3}} & \multicolumn{3}{c|}{\textbf{0.15}} & \multicolumn{3}{c}{\textbf{0.3}} \\ \cline{2-20} 
\multicolumn{1}{c|}{} & \textbf{$\alpha^{\text{ts}}$} & \multicolumn{1}{c}{\textbf{0}} & \multicolumn{1}{c}{\textbf{0.15}} & \multicolumn{1}{c|}{\textbf{0.3}} & \multicolumn{1}{c}{\textbf{0}} & \multicolumn{1}{c}{\textbf{0.15}} & \multicolumn{1}{c|}{\textbf{0.3}} & \multicolumn{1}{c}{\textbf{0}} & \multicolumn{1}{c}{\textbf{0.15}} & \multicolumn{1}{c|}{\textbf{0.3}} & \multicolumn{1}{c}{\textbf{0}} & \multicolumn{1}{c}{\textbf{0.15}} & \multicolumn{1}{c|}{\textbf{0.3}} & \multicolumn{1}{c}{\textbf{0}} & \multicolumn{1}{c}{\textbf{0.15}} & \multicolumn{1}{c|}{\textbf{0.3}} & \multicolumn{1}{c}{\textbf{0}} & \multicolumn{1}{c}{\textbf{0.15}} & \multicolumn{1}{c}{\textbf{0.3}} \\ \midrule
\multicolumn{1}{c|}{\multirow{3}{*}{0.15}} & SAINT & 84.52 & 83.53 & \multicolumn{1}{l|}{82.76} & 83.78 & 83.36 & 82.59 & 84.66 & 83.69 & \multicolumn{1}{l|}{82.26} & 84.03 & 83.41 & 82.42 & 84.47 & 83.64 & \multicolumn{1}{l|}{82.12} & 83.97 & 83.14 & 82.28 \\
\multicolumn{1}{c|}{} & Stab & 79.23 & 77.68 & \multicolumn{1}{l|}{77.34} & 79.57 & 78.25 & 77.01 & 79.51 & 78.23 & \multicolumn{1}{l|}{75.50} & 79.50 & 78.48 & 76.57 & 79.72 & 77.04 & \multicolumn{1}{l|}{74.21} & 78.97 & 77.50 & 75.52 \\ \cline{2-20} 
\multicolumn{1}{c|}{} & Ours & 84.45 & 83.84 & \multicolumn{1}{l|}{83.31} & 84.29 & 83.66 & 83.27 & 84.43 & 83.27 & \multicolumn{1}{l|}{82.39} & 84.15 & 83.23 & 82.59 & 84.56 & 83.64 & \multicolumn{1}{l|}{82.39} & 84.46 & 83.46 & 82.50 \\
\bottomrule
\end{tabular}
}
\label{ssl:tab:blastchar}
\end{table}

\begin{table}[H]
\renewcommand\thetable{B.28}
\centering
\fontsize{6pt}{7pt}\selectfont
\caption{Comparison of averaged test AUC scores (\%) on \texttt{churn} for various $(\alpha^{\text{tr}},\alpha^{\text{ts}},\alpha_y^{\text{tr}})$ combinations.}
\resizebox{0.99\textwidth}{!}{
\begin{tabular}{cl|llllll|llllll|llllll}
\toprule
\multicolumn{2}{c|}{\textbf{Method}} & \multicolumn{6}{c|}{\textbf{MAR}} & \multicolumn{6}{c|}{\textbf{MNAR}} & \multicolumn{6}{c}{\textbf{MCAR}} \\ \midrule
\multicolumn{1}{c|}{\multirow{2}{*}{$\alpha_y^{\text{tr}}$}} & \textbf{$\alpha^{\text{tr}}$} & \multicolumn{3}{c|}{\textbf{0.15}} & \multicolumn{3}{c|}{\textbf{0.3}} & \multicolumn{3}{c|}{\textbf{0.15}} & \multicolumn{3}{c|}{\textbf{0.3}} & \multicolumn{3}{c|}{\textbf{0.15}} & \multicolumn{3}{c}{\textbf{0.3}} \\ \cline{2-20} 
\multicolumn{1}{c|}{} & \textbf{$\alpha^{\text{ts}}$} & \multicolumn{1}{c}{\textbf{0}} & \multicolumn{1}{c}{\textbf{0.15}} & \multicolumn{1}{c|}{\textbf{0.3}} & \multicolumn{1}{c}{\textbf{0}} & \multicolumn{1}{c}{\textbf{0.15}} & \multicolumn{1}{c|}{\textbf{0.3}} & \multicolumn{1}{c}{\textbf{0}} & \multicolumn{1}{c}{\textbf{0.15}} & \multicolumn{1}{c|}{\textbf{0.3}} & \multicolumn{1}{c}{\textbf{0}} & \multicolumn{1}{c}{\textbf{0.15}} & \multicolumn{1}{c|}{\textbf{0.3}} & \multicolumn{1}{c}{\textbf{0}} & \multicolumn{1}{c}{\textbf{0.15}} & \multicolumn{1}{c|}{\textbf{0.3}} & \multicolumn{1}{c}{\textbf{0}} & \multicolumn{1}{c}{\textbf{0.15}} & \multicolumn{1}{c}{\textbf{0.3}} \\ \midrule
\multicolumn{1}{c|}{\multirow{3}{*}{0.15}} & SAINT & 84.69 & 82.41 & \multicolumn{1}{l|}{79.14} & 84.39 & 82.10 & 79.06 & 84.93 & 82.33 & \multicolumn{1}{l|}{79.08} & 84.88 & 82.44 & 79.09 & 84.79 & 82.45 & \multicolumn{1}{l|}{78.67} & 84.25 & 82.10 & 78.60 \\
\multicolumn{1}{c|}{} & Stab & 60.29 & 58.53 & \multicolumn{1}{l|}{57.47} & 53.22 & 51.89 & 51.30 & 58.86 & 57.53 & \multicolumn{1}{l|}{56.85} & 54.87 & 54.22 & 53.67 & 58.21 & 58.06 & \multicolumn{1}{l|}{57.32} & 51.91 & 51.39 & 50.61 \\ \cline{2-20} 
\multicolumn{1}{c|}{} & Ours & 84.66 & 82.48 & \multicolumn{1}{l|}{79.35} & 84.35 & 82.46 & 79.23 & 84.60 & 82.49 & \multicolumn{1}{l|}{79.40} & 84.60 & 82.46 & 79.43 & 84.65 & 82.55 & \multicolumn{1}{l|}{79.37} & 84.25 & 82.21 & 79.05 \\
\bottomrule
\end{tabular}
}
\label{ssl:tab:churn}
\end{table}

\begin{table}[H]
\renewcommand\thetable{B.29}
\centering
\fontsize{6pt}{7pt}\selectfont
\caption{Comparison of averaged test AUC scores (\%) on \texttt{htur2} for various $(\alpha^{\text{tr}},\alpha^{\text{ts}},\alpha_y^{\text{tr}})$ combinations.}
\resizebox{0.99\textwidth}{!}{
\begin{tabular}{cl|llllll|llllll|llllll}
\toprule
\multicolumn{2}{c|}{\textbf{Method}} & \multicolumn{6}{c|}{\textbf{MAR}} & \multicolumn{6}{c|}{\textbf{MNAR}} & \multicolumn{6}{c}{\textbf{MCAR}} \\ \midrule
\multicolumn{1}{c|}{\multirow{2}{*}{$\alpha_y^{\text{tr}}$}} & \textbf{$\alpha^{\text{tr}}$} & \multicolumn{3}{c|}{\textbf{0.15}} & \multicolumn{3}{c|}{\textbf{0.3}} & \multicolumn{3}{c|}{\textbf{0.15}} & \multicolumn{3}{c|}{\textbf{0.3}} & \multicolumn{3}{c|}{\textbf{0.15}} & \multicolumn{3}{c}{\textbf{0.3}} \\ \cline{2-20} 
\multicolumn{1}{c|}{} & \textbf{$\alpha^{\text{ts}}$} & \multicolumn{1}{c}{\textbf{0}} & \multicolumn{1}{c}{\textbf{0.15}} & \multicolumn{1}{c|}{\textbf{0.3}} & \multicolumn{1}{c}{\textbf{0}} & \multicolumn{1}{c}{\textbf{0.15}} & \multicolumn{1}{c|}{\textbf{0.3}} & \multicolumn{1}{c}{\textbf{0}} & \multicolumn{1}{c}{\textbf{0.15}} & \multicolumn{1}{c|}{\textbf{0.3}} & \multicolumn{1}{c}{\textbf{0}} & \multicolumn{1}{c}{\textbf{0.15}} & \multicolumn{1}{c|}{\textbf{0.3}} & \multicolumn{1}{c}{\textbf{0}} & \multicolumn{1}{c}{\textbf{0.15}} & \multicolumn{1}{c|}{\textbf{0.3}} & \multicolumn{1}{c}{\textbf{0}} & \multicolumn{1}{c}{\textbf{0.15}} & \multicolumn{1}{c}{\textbf{0.3}} \\ \midrule
\multicolumn{1}{c|}{\multirow{3}{*}{0.15}} & SAINT & 97.47 & 97.30 & \multicolumn{1}{l|}{96.45} & 97.27 & 97.19 & 96.76 & 97.39 & 97.21 & \multicolumn{1}{l|}{96.82} & 97.45 & 97.04 & 96.69 & 97.45 & 97.24 & \multicolumn{1}{l|}{96.82} & 97.49 & 97.19 & 96.84 \\
\multicolumn{1}{c|}{} & Stab & 96.86 & 96.95 & \multicolumn{1}{l|}{95.95} & 97.20 & 96.80 & 96.23 & 97.30 & 96.84 & \multicolumn{1}{l|}{95.90} & 97.08 & 96.92 & 96.44 & 97.14 & 97.00 & \multicolumn{1}{l|}{96.25} & 96.89 & 97.01 & 96.62 \\ \cline{2-20} 
\multicolumn{1}{c|}{} & Ours & 97.06 & 96.97 & \multicolumn{1}{l|}{96.50} & 96.81 & 96.80 & 96.44 & 96.98 & 96.94 & \multicolumn{1}{l|}{96.82} & 96.82 & 96.84 & 96.67 & 97.08 & 97.04 & \multicolumn{1}{l|}{96.65} & 96.91 & 96.96 & 96.54 \\
\bottomrule
\end{tabular}
}
\label{ssl:tab:htur2}
\end{table}

\begin{table}[H]
\renewcommand\thetable{B.30}
\centering
\fontsize{6pt}{7pt}\selectfont
\caption{Comparison of averaged test AUC scores (\%) on \texttt{qsar\_bio} for various $(\alpha^{\text{tr}},\alpha^{\text{ts}},\alpha_y^{\text{tr}})$ combinations.}
\resizebox{0.99\textwidth}{!}{
%
}
\label{ssl:tab:qsar_bio}
\end{table}

\begin{table}[H]
\renewcommand\thetable{B.31}
\centering
\fontsize{6pt}{7pt}\selectfont
\caption{Comparison of averaged test AUC scores (\%) on \texttt{shoppers} for various $(\alpha^{\text{tr}},\alpha^{\text{ts}},\alpha_y^{\text{tr}})$ combinations.}
\resizebox{0.99\textwidth}{!}{
%
}
\label{ssl:tab:shoppers}
\end{table}

\begin{table}[H]
\renewcommand\thetable{B.32}
\centering
\fontsize{6pt}{7pt}\selectfont
\caption{Comparison of averaged test AUC scores (\%) on \texttt{spambase} for various $(\alpha^{\text{tr}},\alpha^{\text{ts}},\alpha_y^{\text{tr}})$ combinations.}
\resizebox{0.99\textwidth}{!}{
%
}
\label{ssl:tab:spambase}
\end{table}

\newpage
\subsection{C. Ablation studies}
\label{sec:appC}
\paragraph{Masking Ratio Selection} 
Positive values of $r$ generally lead to improved performance of our method. 
However, as $r$ increases beyond a certain threshold, this positive effect diminishes, and performance begins to deteriorate. 
These results align with our conjecture: 
the use of additional masking operations acts as an implicit regularizer, thereby improving prediction performance when $r$ is set to an appropriate value.
However, if $r$ is too large, it can cause a substantial mismatch between the simulated and actual test-time missingness distributions, which undermines the theoretical justification for using the modified MI conditions and consequently degrades performance.
As a result, the experiment indicates that our method is not sensitive to the value of $r$ in practice, as long as it is chosen reasonably. 

%

\begingroup
\renewcommand{\arraystretch}{0.99}
\begin{table}[H]
\renewcommand\thetable{C.1}
\caption{Averaged results of test AUC scores with various values of $r$.}
\resizebox{0.99\textwidth}{!}{%
\begin{tabular}{l|ccccc|ccccc|ccccc|ccccc}
\toprule
$(\alpha^{\text{tr}},\alpha^{\text{ts}})$ & \multicolumn{5}{c|}{(0.15,   0.15)} & \multicolumn{5}{c|}{(0.15, 0.3)} & \multicolumn{5}{c|}{(0.3, 0.15)} & \multicolumn{5}{c}{(0.3, 0.3)} \\ \midrule
$r$ & 0 & 0.1 & 0.2 & 0.3 & 0.4 & 0 & 0.1 & 0.2 & 0.3 & 0.4 & 0 & 0.1 & 0.2 & 0.3 & 0.4 & 0 & 0.1 & 0.2 & 0.3 & 0.4 \\ \midrule
adult & 90.67 & 91.03 & 91.14 & 91.13 & 91.05 & 88.70 & 89.57 & 89.74 & 89.77 & 89.70 & 90.58 & 90.93 & 90.93 & 90.88 & 90.91 & 89.00 & 89.47 & 89.57 & 89.56 & 89.56 \\
htru2 & 97.29 & 97.45 & 97.45 & 97.46 & 97.27 & 96.56 & 97.14 & 97.11 & 97.19 & 97.15 & 97.31 & 97.19 & 97.14 & 97.14 & 97.00 & 96.77 & 96.82 & 96.85 & 96.92 & 96.81 \\
qsar\_bio & 88.28 & 89.39 & 89.73 & 90.04 & 89.89 & 85.16 & 87.42 & 88.42 & 89.33 & 89.21 & 89.30 & 89.94 & 89.67 & 89.90 & 89.04 & 88.05 & 88.92 & 88.85 & 89.25 & 88.47 \\ \bottomrule
\end{tabular}}
\end{table} 
\endgroup

\paragraph{Loss Term Analysis} 
With respect to the hyperparameters $\lambda_1$ and $\lambda_2$, which control the weights of the proposed loss components $\mathcal{L}_2^{\text{MI}}$ and $\mathcal{L}_3^{\text{MI}}$, we observe that the method performs best when both values are set to be positive. 
This indicates that these MI-based regularization terms contribute meaningfully to model learning. 
We note that $\mathcal{L}_2^{\text{MI}}$ and $\mathcal{L}_3^{\text{MI}}$ appear to play a similar role at first glance, as both encourage the prediction model to maintain correct predictions even under additional masking operations.
However, the third term, $\mathcal{L}_3^{\text{MI}}$, functions as a loss only for inputs on which the model exhibits high confidence. 
As a result, $\mathcal{L}_3^{\text{MI}}$ places greater loss weight on high-confidence examples, effectively propagating reliable information to more challenging instances and thereby enhancing the overall learning process.

\renewcommand{\arraystretch}{0.9}
\begin{table}[h]
\centering
\caption{Averaged results of test AUC scores with various values of $\lambda_1$.}
\fontsize{8pt}{9.6pt}\selectfont
\resizebox{0.85\textwidth}{!}{%
\begin{tabular}{l|cccccc|cccccc}
\toprule
$(\alpha^{\text{tr}},\alpha^{\text{ts}})$ & \multicolumn{6}{c|}{(0.15,   0.15)} & \multicolumn{6}{c}{(0.15, 0.3)} \\ \midrule
$\lambda_1$ & 0 & 1 & 5 & 10 & 15 & 20 & 0 & 1 & 5 & 10 & 15 & 20 \\ \midrule
adult & 90.25 & 90.88 & 91.14 & 91.14 & 91.14 & 91.18 & 88.17 & 89.31 & 89.71 & 89.73 & 89.74 & 89.83 \\
htru2 & 96.88 & 97.24 & 97.48 & 97.47 & 97.45 & 97.43 & 96.20 & 96.86 & 97.23 & 97.16 & 97.11 & 97.13 \\
qsar\_bio & 87.64 & 90.02 & 89.74 & 89.79 & 89.73 & 89.74 & 83.23 & 88.61 & 88.74 & 88.42 & 88.42 & 88.59 \\ \bottomrule
\end{tabular}}
\par\vspace{0.3cm}
\resizebox{0.85\textwidth}{!}{%
\begin{tabular}{l|cccccc|cccccc}
\toprule
$(\alpha^{\text{tr}},\alpha^{\text{ts}})$ & \multicolumn{6}{c|}{(0.3,   0.15)} & \multicolumn{6}{c}{(0.3, 0.3)} \\ \midrule
$\lambda_1$ & 0 & 1 & 5 & 10 & 15 & 20 & 0 & 1 & 5 & 10 & 15 & 20 \\ \midrule
adult & 90.43 & 90.67 & 90.91 & 90.94 & 90.93 & 90.93 & 88.76 & 89.30 & 89.56 & 89.57 & 89.57 & 89.56 \\
htru2 & 96.76 & 97.05 & 97.16 & 97.12 & 97.14 & 97.15 & 96.36 & 96.80 & 96.89 & 96.84 & 96.85 & 96.86 \\
qsar\_bio & 85.90 & 89.17 & 89.86 & 89.76 & 89.67 & 89.80 & 83.87 & 88.30 & 89.18 & 88.92 & 88.85 & 89.17 \\ \bottomrule
\end{tabular}}

\end{table}

\renewcommand{\arraystretch}{0.9}
\begin{table}[h]
\centering
\caption{Averaged results of test AUC scores with various values of $\lambda_2$.}
\fontsize{8pt}{9.6pt}\selectfont
\resizebox{0.85\textwidth}{!}{%
\begin{tabular}{l|cccccc|cccccc}
\toprule
$(\alpha^{\text{tr}},\alpha^{\text{ts}})$ & \multicolumn{6}{c|}{(0.15,   0.15)} & \multicolumn{6}{c}{(0.15, 0.3)} \\ \midrule
$\lambda_2$ & 0 & 1 & 5 & 10 & 15 & 20 & 0 & 1 & 5 & 10 & 15 & 20 \\ \midrule
adult & 91.10 & 91.10 & 91.10 & 91.11 & 91.14 & 91.16 & 89.73 & 89.74 & 89.74 & 89.76 & 89.74 & 89.78 \\
htru2 & 97.39 & 97.43 & 97.40 & 97.41 & 97.45 & 97.43 & 97.02 & 97.07 & 97.05 & 97.07 & 97.11 & 97.12 \\
qsar\_bio & 89.69 & 89.68 & 89.72 & 89.70 & 89.73 & 89.79 & 88.18 & 88.29 & 88.36 & 88.47 & 88.42 & 88.29 \\ \bottomrule
\end{tabular}}
\par\vspace{0.3cm}
\resizebox{0.85\textwidth}{!}{%
\begin{tabular}{l|cccccc|cccccc}
\midrule
$(\alpha^{\text{tr}},\alpha^{\text{ts}})$ & \multicolumn{6}{c|}{(0.3,   0.15)} & \multicolumn{6}{c}{(0.3, 0.3)} \\ \midrule
$\lambda_2$ & 0 & 1 & 5 & 10 & 15 & 20 & 0 & 1 & 5 & 10 & 15 & 20 \\ \midrule
adult & 90.96 & 90.96 & 90.97 & 90.98 & 90.93 & 90.93 & 89.57 & 89.57 & 89.59 & 89.60 & 89.57 & 89.57 \\
htru2 & 97.16 & 97.16 & 97.15 & 97.15 & 97.14 & 97.12 & 96.85 & 96.85 & 96.86 & 96.86 & 96.85 & 96.84 \\
qsar\_bio & 89.66 & 89.65 & 89.71 & 89.80 & 89.67 & 89.73 & 88.78 & 88.72 & 88.80 & 88.97 & 88.85 & 89.05 \\ \bottomrule
\end{tabular}}
\end{table}

\newpage
\paragraph{Confidence Threshold Selection} 
Lastly, we find that the method is relatively insensitive to the choice of the temperature parameter $\tau$ across the range of values we considered. 
This robustness indicates that the effect of $\mathcal{L}_3^{\text{MI}}$--namely, propagating reliable information to more challenging instances--remains effective as long as $\tau$ is set within a reasonable range. 
In addition, our method does not require fine-tuning of $\tau$ to achieve strong performance, which improves usability in practical settings.

\begin{table}[H]
    \centering
    \caption{Averaged results of test AUC scores with various values of $\tau$.}
    \resizebox{0.95\textwidth}{!}{%
    \begin{tabular}{l|cccc|cccc|cccc|cccc}
    \toprule
    $(\alpha^{\bold{tr}}, \alpha^\bold{ts})$ & \multicolumn{4}{c|}{(0.15,   0.15)} & \multicolumn{4}{c|}{(0.15, 0.3)} & \multicolumn{4}{c|}{(0.3, 0.15)} & \multicolumn{4}{c}{(0.3, 0.3)} \\ \midrule
    $\tau$ & 0.8 & 0.9 & 0.95 & 0.99 & 0.8 & 0.9 & 0.95 & 0.99 & 0.8 & 0.9 & 0.95 & 0.99 & 0.8 & 0.9 & 0.95 & 0.99 \\ \midrule
    adult & 91.20 & 91.13 & 91.14 & 91.18 & 89.81 & 89.81 & 89.74 & 89.82 & 90.98 & 90.98 & 90.93 & 90.95 & 89.65 & 89.63 & 89.57 & 89.57 \\
    htru2 & 97.49 & 97.42 & 97.45 & 97.44 & 97.22 & 97.13 & 97.11 & 97.18 & 97.20 & 97.17 & 97.14 & 97.20 & 96.92 & 96.88 & 96.85 & 96.93 \\
    qsar\_bio & 89.65 & 89.79 & 89.73 & 89.69 & 88.66 & 88.72 & 88.42 & 88.26 & 89.57 & 89.63 & 89.67 & 89.66 & 88.94 & 89.02 & 88.85 & 88.77 \\ \bottomrule
    \end{tabular}}
\end{table}

\paragraph{Running Time Analysis} 
Although our method requires two forward passes per update--one for the original training data and another for the additionally masked version--it maintains faster running time compared to other deep learning–based methods, in a case exceeding a three times speedup.
This efficiency is largely due to the fact that our approach does not employ any auxiliary architectures such as decoders, while other competitors do.
Therefore, this result highlights our method as an efficient learning framework in terms of both memory usage and computational cost.

\begin{table}[h]
	\centering
    \caption{Comparison of running times. All results are reported as the average running time per epoch across the benchmark datasets, rescaled such that the running time of our method is normalized to one.}
    \small
    \begin{tabular}{l|cccc}
    \toprule
    Method      & MIRRAMS & SwitchTab & SAINT  & TabTransformer \\ \midrule
    Time(ratio) & 1       & 1.3365    & 2.5837 & 3.7881         \\ \bottomrule
    \end{tabular}
\end{table}

\end{document}